\documentclass[10pt,twocolumn,letterpaper]{article}

\usepackage[accsupp]{axessibility}  

\usepackage[pagenumbers]{cvpr} 

\definecolor{cvprblue}{rgb}{0.21,0.49,0.74}
\usepackage[pagebackref,breaklinks,colorlinks,allcolors=cvprblue]{hyperref}

\usepackage[dvipsnames]{xcolor}
\definecolor{mygreen}{HTML}{009700} 
\usepackage{url}
\usepackage{algorithm}
\usepackage{algpseudocode}
\usepackage{booktabs}
\usepackage{multirow}      
\usepackage{makecell}  
\usepackage{amsmath,amssymb}
\usepackage{amsthm}        
\usepackage{hyperref}      
\usepackage{courier}       
\newtheorem{theorem}{Theorem}[section]
\newtheorem{proposition}[theorem]{Proposition}
\theoremstyle{remark}

\usepackage{xspace}
\usepackage{graphicx}
\usepackage{xcolor}
\usepackage{colortbl}
\usepackage{wrapfig}
\usepackage{graphicx}
\usepackage{subcaption}
\usepackage{xcolor}

\hypersetup{
    colorlinks,
    citecolor=cvprblue
}

\def\confName{CVPR}
\def\confYear{2026}

\title{ZOO-Prune: Training-Free Token Pruning via \\ Zeroth-Order Gradient Estimation in Vision-Language Models}

\author{
\textbf{Youngeun Kim}\textsuperscript{1}\textsuperscript{$\ast$ $\dagger$} \
\textbf{Youjia Zhang}\textsuperscript{2}\textsuperscript{$\ast$} \
\textbf{Huiling Liu}\textsuperscript{2} \
\textbf{Aecheon Jung}\textsuperscript{2} \
\textbf{Sunwoo Lee}\textsuperscript{3} \
\textbf{Sungeun Hong}\textsuperscript{2}\textsuperscript{$\ddagger$} \\ [0.2cm]
\textsuperscript{1}Amazon \quad \textsuperscript {2}Sungkyunkwan University \quad \textsuperscript {3}Inha University\quad
}

\newcommand{\ie}{\textit{i}.\textit{e}., }
\newcommand{\eg}{\textit{e}.\textit{g}., }

\newcommand{\ours}{\textit{ZOO-Prune}\xspace}

\begin{document}
\maketitle

\begingroup
\renewcommand{\thefootnote}{}
\footnotetext{$\ast$ Equal contribution.}
\footnotetext{$\dagger$ This work was completed prior to the author joining Amazon.}
\footnotetext{$\ddagger$ Corresponding author: Sungeun Hong (csehong@skku.edu)}
\endgroup

\begin{abstract}
Large Vision–Language Models (VLMs) enable strong multimodal reasoning but incur heavy inference costs from redundant visual tokens. Token pruning alleviates this issue, yet existing approaches face limitations. Attention-based methods rely on raw attention scores, which are often unstable across layers and heads and can lead to redundant selections. Diversity-based methods improve robustness by selecting tokens far apart in feature space, but risk dropping regions needed for accurate prediction.
We propose \ours, a training-free framework built on the intuition that highly sensitive tokens have a stronger influence on the model's output and capture complementary visual cues rather than redundant ones. To achieve this, we estimate token sensitivity using zeroth-order perturbations at the lightweight projection layer. This measures how small random perturbations affect the projected features and enables efficient approximation of each token’s influence without backpropagation.
Extensive experiments across multiple VLMs and benchmarks show that \ours consistently outperforms prior methods while pruning up to 94.4\% of tokens without sacrificing accuracy. Our method also improves efficiency, reaching up to 2.30$\times$ faster end-to-end inference compared to the baseline. Code is available at \hypersetup{urlcolor=magenta}\url{https://aim-skku.github.io/ZOO-Prune}.

\end{abstract}


\begin{figure*}[t]
\begin{center}
\centering
\vspace{-4mm}
\def\arraystretch{0.5}
\begin{tabular}{@{\hskip 0.01\linewidth}c@{\hskip 0.03\linewidth}c@{}c}
\hspace{-4mm}
\includegraphics[width=0.98\linewidth]
{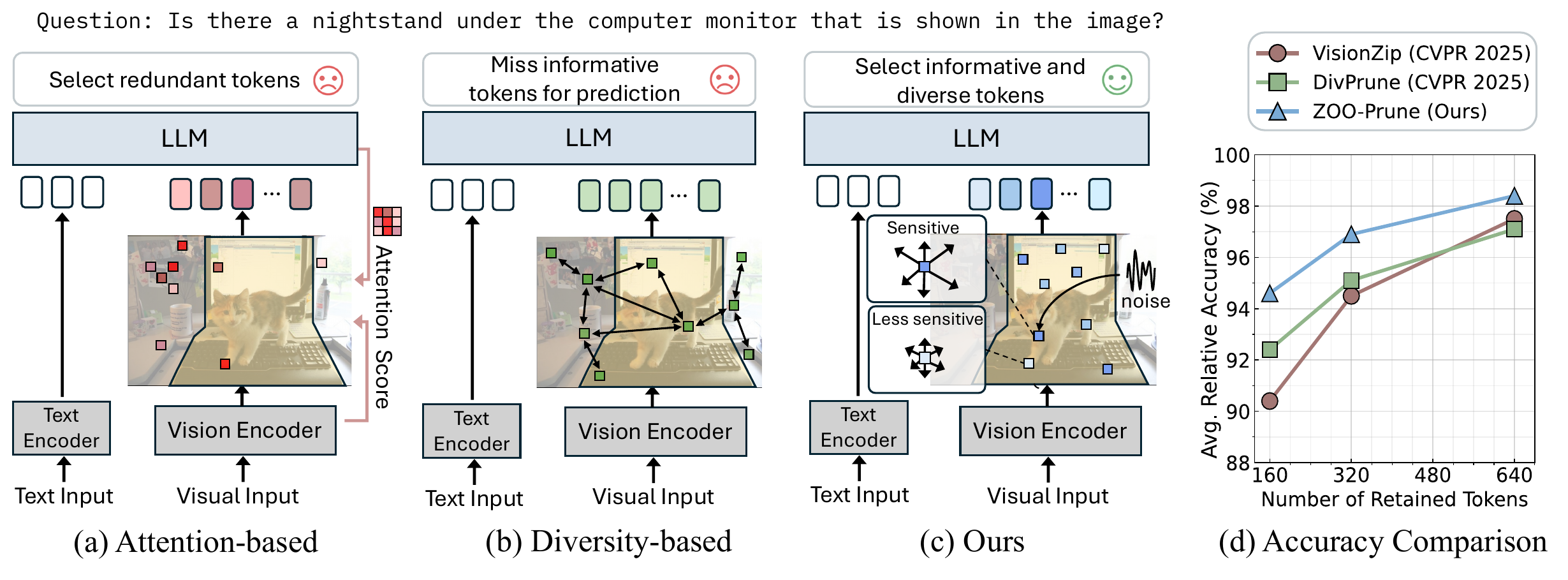} 
\end{tabular}
\end{center}
\vspace{-8mm}
\caption{
Illustration of training-free VLM token pruning methods. (a) Attention-based methods select tokens using attention scores, but often retain redundant tokens. (b) Diversity-based methods select tokens with different features to maximize coverage but may lose tokens located in semantically relevant regions (\eg around the monitor, highlighted in yellow). (c) Our method employs zeroth-order gradient estimation to quantify token sensitivity and integrates these scores into a diversity objective.
(d) Accuracy comparison with LLaVA-NeXT-7B across 9 benchmarks, showing that ours outperforms both VisionZip (attention-based) and DivPrune (diversity-based). }
\vspace{-3mm}
\label{fig:intro:concept}
\end{figure*}

\vspace{-3mm}
\section{Introduction}
Large Vision–Language Models (VLMs)~\citep{qwen,LLaVA-NeXT,internvl3} achieve strong multimodal understanding but at a substantial computational cost. A major contributor to this cost is the large number of visual tokens produced by modern vision backbones. For example, the vision encoder in LLaVA-1.5~\citep{LLaVA-1.5} generates up to 576 tokens for a single image, whereas the text side often contains only a few tokens, such as \textit{“Describe this image in a short sentence”}. This imbalance leads to high inference latency and memory overhead~\citep{FastV,Divprune}. To address this issue, token pruning has emerged as a practical solution that selectively removes less informative visual tokens at inference time. Prior work has shown that aggressive token reduction can yield large speedups with only modest drops in accuracy~\citep{sparsevlm,visionzip,hired}.

Recent research on VLM token pruning has increasingly focused on training-free schemes. Unlike works that require calibration data or fine-tuning~\citep{GQA,Tokenpacker,VTW}, training-free approaches prune tokens directly at inference. These methods can be broadly categorized into two groups. (1) Attention-based approaches score tokens using attention magnitudes; however, attention often concentrates on background regions~\citep{darcet2023vision} and tends to retain redundant tokens with overlapping content~\citep{visionzip,Llava-prumerge}. For example, in an image of a laptop on a desk (Fig.~\ref{fig:intro:concept}), attention-based pruning may keep many redundant background tokens while missing tokens near the monitor that are important for answering questions. (2) Diversity-based approaches~\citep{Divprune,balanced, pei2025greedyprune} select tokens by maximizing feature diversity, measuring pairwise distances between token embeddings. However, because these methods prioritize diversity uniformly across all tokens without explicitly considering task-relevant cues, they may discard tokens from visually salient regions. In the same laptop example, diversity-based pruning may select feature-diverse tokens, but not necessarily those near the monitor, which are crucial for reasoning about nearby objects.


Given these limitations, selecting a smaller subset of tokens while preserving visual information without degrading task performance remains a challenging problem.
To tackle this, we investigate a novel metric for scoring each token, explicitly considering token sensitivity (\ie the degree to which small perturbations to a token change the model’s output). 
Prior work has shown that attention weights do not necessarily correlate with a token’s actual impact on the model output~\citep{brunner2019identifiability,wu2024token}, motivating the use of sensitivity-based measures that directly quantify output changes under token perturbations~\citep{chefer2021transformer}.
Unfortunately, directly computing token sensitivities via gradients is costly. It also requires ground-truth outputs to define a supervised loss, which is unavailable during inference-time pruning. These factors motivate us to use zeroth-order gradient estimation~\citep{nesterov2017random}, which quantifies how perturbations to input tokens affect the model output using only forward passes, avoiding backpropagation.

However, a naive application of zeroth-order estimation would require additional forward passes through the vision encoder, incurring substantial computational overhead. 
We therefore empirically examined whether sensitivity rankings obtained at a lightweight intermediate stage could serve as a reliable proxy for those derived from the full vision encoder. Intuitively, the projection layer is a natural choice since tokens here already capture high-level semantics from the vision encoder and are directly aligned with the language model~\citep{cha2024honeybee, wang2025pargo}. Our analysis revealed strong alignment between rankings computed from the vision encoder and those from the projection layer (Section~\ref{subsec:observation}).

Motivated by this finding, we propose a training-free VLM token pruning method named \textit{\ours (ZerOth-Order gradient estimation for token pruning)}. Our method measures each token’s influence by injecting Gaussian noise at a lightweight projection layer and estimating the resulting gradient norms, which we define as sensitivity. To reduce redundancy while retaining informative tokens, we introduce sensitivity-aware diversity selection, which prioritizes tokens with high sensitivity and ensures sufficient feature diversity, inspired by~\citep{Divprune}. By jointly considering sensitivity and diversity, \ours produces pruned token subsets that preserve task-relevant information and enable effective compression even under aggressive pruning regimes (Fig.~\ref{fig:intro:concept}).
Our main contributions are as follows:
\begin{itemize}
\setlength{\itemsep}{0.5pt}    
\item We propose \ours, a training-free pruning framework that unifies sensitivity and diversity, ensuring that pruned tokens are not only highly informative but also complementary, overcoming the limitations of attention-only or diversity-only methods.
\item We introduce a zeroth-order sensitivity estimator at the projection layer, which provides stable token importance rankings with lightweight forward computations, eliminating the need for backpropagation or costly full-encoder passes.
\item We demonstrate through extensive experiments on multiple VLMs and benchmarks that \ours delivers superior accuracy–efficiency trade-offs. It retains up to 94.4\% fewer tokens while maintaining accuracy and significantly reducing inference cost, achieving up to {2.30$\times$} faster end-to-end inference compared to the baseline.
\end{itemize}

\section{Related Work}
\subsection{Vision-Language Models}
Large Multimodal Models (LMMs), particularly Vision-Language Models (VLMs)~\citep{LLaVA, instructblip, internvl, LLaVA-NeXT,qwen}, have demonstrated remarkable capabilities in multimodal reasoning and understanding~\citep{wangprimt,zhang2025cat,cho2025ra}. Pioneering architectures like LLaVA~\citep{LLaVA} established a successful paradigm by aligning a pre-trained vision encoder (\eg CLIP ViT~\citep{CLIP}) with an instruction-tuned LLM through a simple projection layer. This design, further refined in subsequent works such as LLaVA family~\citep {LLaVA-1.5,LLaVA-NeXT}, InternVL series~\citep{internvl,internvl2,internvl3, internvl3_5}, and Qwen-VL~\citep{Qwen2.5}, enables strong visual understanding but introduces a significant computational challenge. These models typically encode a single image into hundreds of visual tokens, leading to substantial inference overhead. The computational burden is exacerbated when handling higher resolutions; LLaVA~\citep{LLaVA, LLaVA-1.5, LLaVA-NeXT} typically encodes 336$\times$336 images into 576 tokens, and up to 2880 tokens at 672$\times$672 resolution. The inherent redundancy within these extensive visual token sequences has motivated research into visual token pruning for efficient VLMs inference.

\subsection{Visual Token Pruning for VLMs}
Visual token pruning methods are proposed to reduce the inference complexity of large VLMs by removing redundant visual representations~\citep{GQA,Tokenpacker,VTW}. Existing methods can be divided into two categories: those that require fine-tuning or calibration, and those that are entirely training-free. The first category relies on additional data or adaptation to guide token reduction. CrossGET~\citep{crossget} and MADTP~\citep{madtp} introduce modality-specific tokens to align cross-modal features and drive token selection, while DeCo~\citep{deco} employs adaptive pooling to decouple token compression from semantic abstraction at the patch level. VTW~\citep{VTW} removes all vision tokens after a specific layer, identified using a small calibration set and a KL-divergence criterion. 
Several approaches have been proposed that target the attention mechanism, motivated by its widespread applicability across diverse tasks and architectures~\cite{zhang2022spatio,wang2024hyper,jung2024scale,ma2023llm}.
FitPrune~\citep{FitPrune} reduces visual tokens in the multi-head attention of each layer via binary search, guided by attention statistics collected from inference or calibration examples. While effective, these methods require calibration data and model-specific adaptation, limiting their flexibility.

In line with efficiency-oriented works widely adopted across diverse domains~\citep{zhang2025memory,zhang2025backpropagation,kim2025task,shin2025dynamic,cho2025discord}, training-free pruning avoids retraining and offers plug-and-play acceleration. Attention-based methods estimate token importance directly from attention magnitudes. FastV~\citep{FastV}, LLaVA-PruMerge~\citep{Llava-prumerge}, ToMe~\citep{tome}, and VisionZip~\citep{visionzip} remove or merge tokens based on early attention maps, while SparseVLM~\citep{sparsevlm} and PyramidDrop~\citep{pdrop} leverage question-driven cross-attention to induce dynamic sparsity. However, attention scores are often unstable and may retain semantically redundant tokens~\citep{Divprune,VTW}, which limits performance under aggressive pruning. Another direction emphasizes feature diversity. DivPrune~\citep{Divprune} formulates token selection as a max-min diversity problem solved via greedy farthest-point sampling, reducing redundancy and preserving robustness at high pruning ratios. Yet by treating all tokens equally, it can overlook semantically critical regions and drop task-relevant information.
To overcome these limitations, we propose a zeroth-order sensitivity estimator that quantifies each token’s effect on the output without backpropagation. We further combine this sensitivity measure with a diversity-based selection strategy so that pruning preserves the most influential tokens while still capturing complementary visual content.
\section{Methodology}
\label{sec:method}


\subsection{Zeroth-Order Gradient Estimation}
\label{subsec:RGE}

Zeroth-order (ZO) optimization offers a gradient-free alternative to first-order methods by using only forward queries of a function. It is particularly useful when exact gradient computation is infeasible, such as in black-box optimization~\citep{sawadanatural}, adversarial attacks~\citep{chen2017zoo}, or efficient fine-tuning of large models~\citep{park2025zip,zhang2024revisiting}. 
In our setting, computing exact gradients would require full backpropagation through a large-scale LLM, which is prohibitively expensive at inference time.
By relying solely on forward evaluations, ZO methods circumvent the need for backpropagation, thereby reducing memory costs and enabling applications to complex or non-differentiable modules.

 A widely used estimator is the \emph{randomized gradient estimator} (RGE)~\citep{duchi2015optimal,nesterov2017random}, which approximates gradients by finite differences along random directions.  
Given a function \(f:\mathbb{R}^{d_1}\to\mathbb{R}^{d_2}\), the central-difference RGE with $m$ queries is:
\begin{gather}
\widehat{\nabla} f(x) \;=\;
\frac{1}{m}\sum_{j=1}^m
\frac{f(x+h u_j) - f(x-h u_j)}{2h}\, u_j,\\
\quad u_j \sim \mathcal{N}(0,I_d),
\end{gather}
where \(h>\!0\) is a small step size.

\subsection{Projector as a Proxy for Visual Sensitivity}
\label{subsec:observation}

\begin{figure}[t]{
  \centering
  \includegraphics[width=\linewidth]{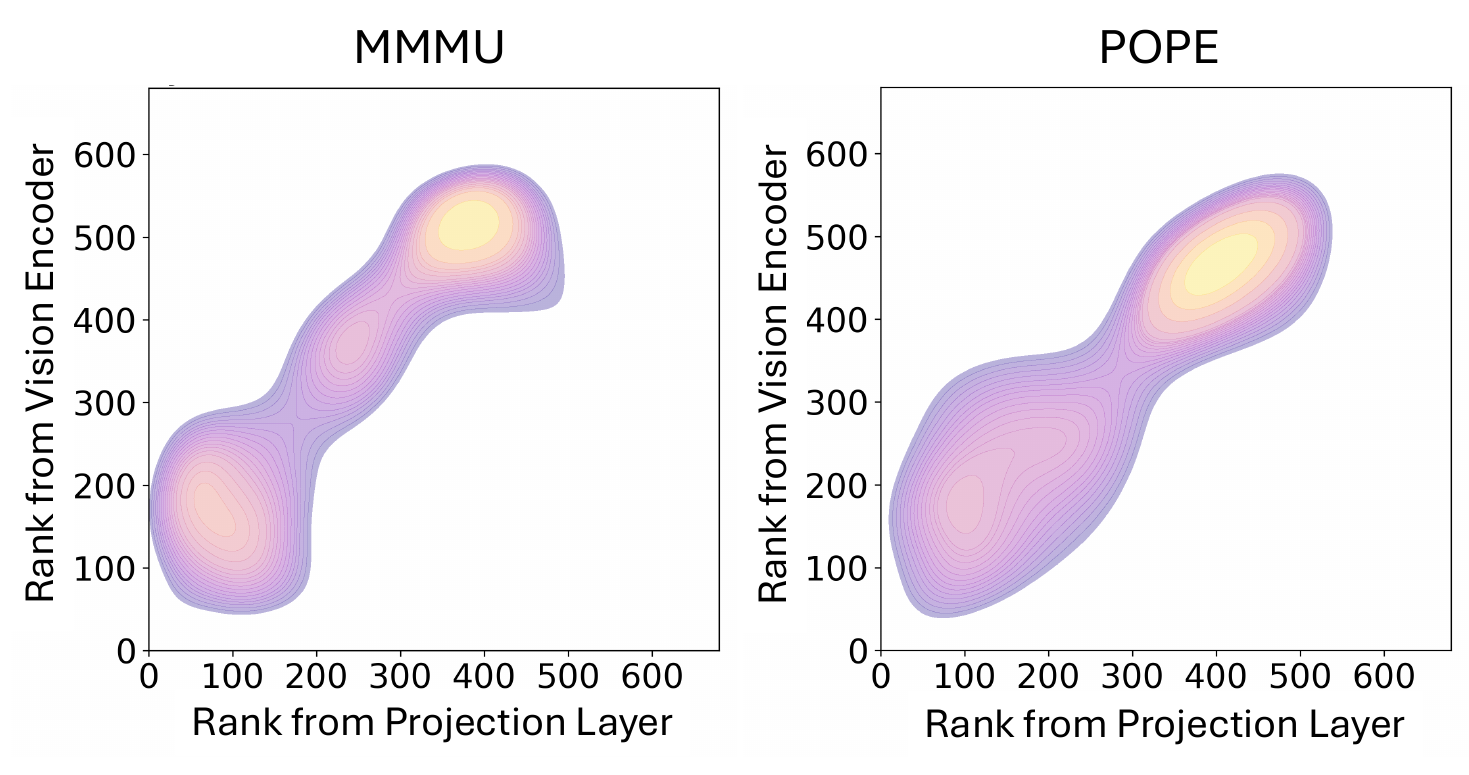}
  \vspace{-5mm} 
  \caption{
Kernel density estimate (KDE) of Spearman rank correlations between token-importance rankings from the \textit{Vision encoder} and the \textit{Projection layer} on the MMMU and POPE datasets. Each dataset shows Spearman correlation of 0.55 and 0.49, respectively. 
Detailed setting is described in Appendix A.
}
\vspace{-5mm}
  \label{fig:method:kde_spear}
  }
\end{figure}

\begin{figure*}[!t]
\begin{center}
\centering
\def\arraystretch{0.5}
\begin{tabular}{@{\hskip 0.01\linewidth}c@{\hskip 0.03\linewidth}c@{}c}
\includegraphics[width=0.88\linewidth]{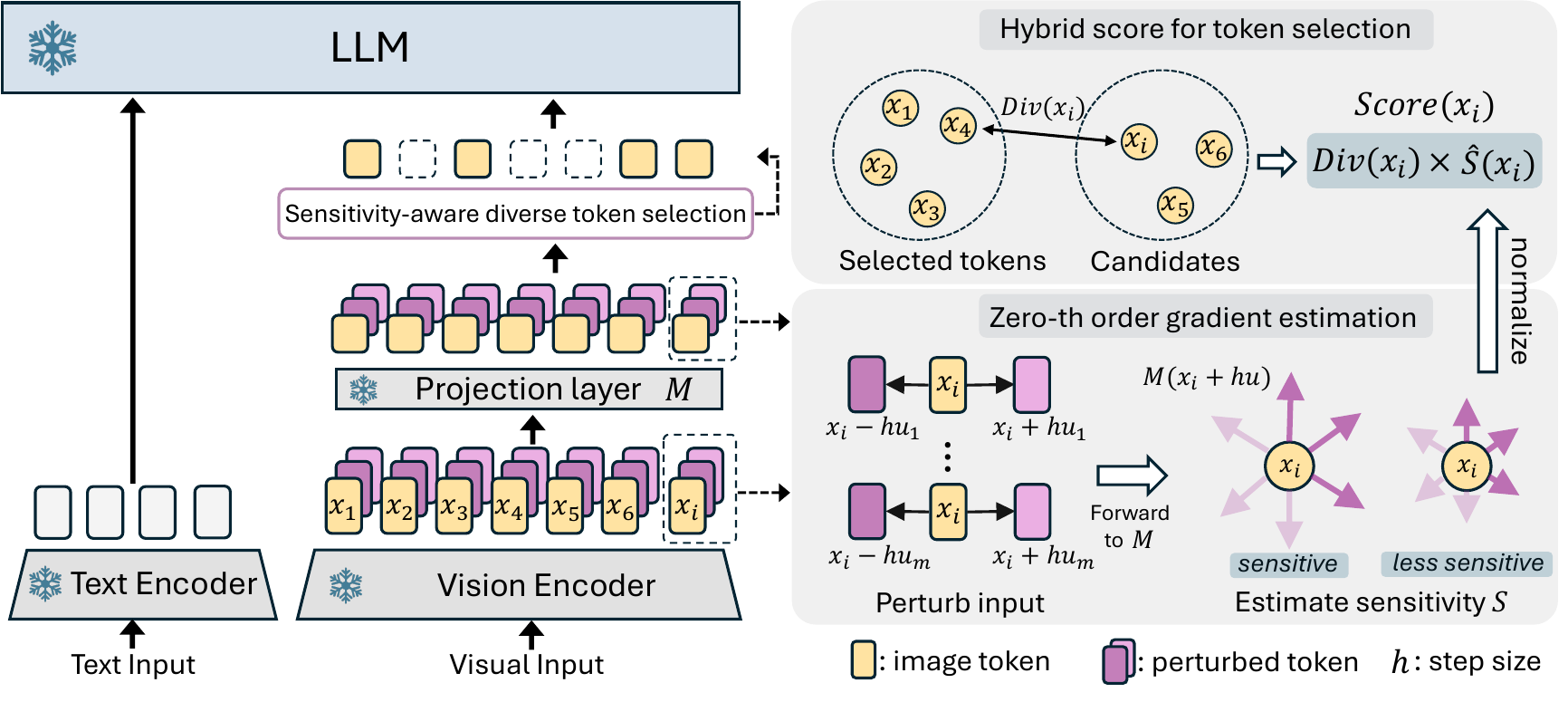} 
\end{tabular}
\end{center}
\vspace{-4mm}
\caption{ 
Overview of \ours. Given visual tokens from the vision encoder, we estimate token sensitivity via zeroth-order gradient approximation at the projection layer by adding Gaussian perturbations (\ie $x_i \pm hu_j$). The resulting sensitivity scores are integrated with a diversity objective to form a hybrid score, guiding the selection. The selected subset is then passed to the LLM together with the text input, enabling efficient multimodal reasoning with reduced computation.}
\label{fig:method:overall_figure}
\end{figure*}

Original zeroth-order methods aim to reconstruct the full gradient vector, but in our setting, we only require a \emph{relative ranking} of token importance. A naive approach would apply RGE directly to the vision encoder, incurring substantial computational overhead due to the additional end-to-end forward passes. For instance, suppose an image is tokenized into $n = 500$ visual tokens with $m = 64$ random perturbation directions per token. Since RGE requires two forward passes per direction, the total cost scales as $2n m$ forward passes, amounting to $\sim6.4 \times 10^6$~GFLOPs, which is clearly prohibitive.

To reduce this cost, we empirically evaluated whether sensitivity rankings obtained at a lightweight intermediate stage (\ie projection layer) could serve as a reliable proxy for those derived from the full vision encoder. In Fig.~\ref{fig:method:kde_spear}, we rank token sensitivities based on RGE and visualize the Spearman rank correlations across two datasets. Our analysis shows a feasible alignment between token importance rankings computed from the vision encoder outputs and those obtained from the projection layer.
Beyond empirical evidence, the projection layer can be seen as a {modality-aligning bottleneck}. It consolidates high-level semantic information from the vision encoder and maps it into the language embedding space, naturally emphasizing tokens that are important for downstream predictions. Since token pruning only requires relative importance rather than exact gradients, this layer provides a compact, semantically meaningful proxy that preserves token sensitivity rankings.

Motivated by this finding, we compute token sensitivities at the projection layer, which contains only a few layers and therefore introduces negligible additional cost during inference. By relying solely on forward queries, this approach avoids the expensive backpropagation required by first-order methods, making zeroth-order sensitivity estimation a practical and efficient tool for token-level analysis in large-scale VLMs.

\subsection{ZOO-Prune}
\label{subsec:ZOO-Prune}

Fig.~\ref{fig:method:overall_figure} illustrates the overall \textit{\ours (ZerOth-Order gradient estimation for token pruning)} framework. 
Given the outputs of the vision encoder, our method first computes token-level sensitivity using zeroth-order gradient estimation and then uses these scores in a diversity-aware selection procedure to produce a subset of tokens that is passed to the LLM. 
Mathematically, let $X \in \mathbb{R}^{N_v \times d_v}$ denote $N_v$ vision tokens with $d_v$ dimension, and
$Z = M(X) \in \mathbb{R}^{N_v \times d_l}$
the projected embeddings through the multimodal projection layer $M: \mathbb{R}^{d_v} \to \mathbb{R}^{d_l}$.  
We sample $m$ random perturbation directions $\{u_j\}_{j=1}^m$, where $u_j \sim \mathcal{N}(0,I_{d_v})$ normalized to unit norm.  
For each token $i$, we measure the symmetric finite-difference response:
\begin{equation}\label{eq:delta_def}
    \delta_{i,j} \;=\; \frac{M(x_i + h u_j) - M(x_i - h u_j)}{2h},
\end{equation}
where $x_i \in \mathbb{R}^{d_v}$ is the $i$-th vision token.  
We define the \emph{token sensitivity} of token $i$ as the average response magnitude:
\begin{equation}\label{eq:sens}
    S(i) \;=\; \frac{1}{m} \sum_{j=1}^m \|\delta_{i,j}\|_2.
\end{equation}
This metric estimates the \emph{approximated mean sensitivity} of each token shown in Proposition~\ref{prop:sensitivity}.

\begin{proposition}[Approximated Mean Sensitivity]\label{prop:sensitivity}
Let $M:\!\mathbb{R}^n \!\to\! \mathbb{R}^m$ be differentiable at $x \!\in\! \mathbb{R}^n$ with Jacobian $J(x) \!=\! \nabla M(x)$. 
Let $u \sim \mathcal{N}(0, I_n)$ be an isotropic Gaussian perturbation and $h \!>\! 0$ a small step size. 
Define the finite-difference sensitivity
$S(x)\!=\!\mathbb{E}_u\left[\left\|\frac{M(x+hu) - M(x-hu)}{2h}\right\|_2\right].$
Then, for sufficiently small $h$:
\begin{equation}
S(x) = \mathbb{E}_u \big[ \| J(x) u \|_2 \big] + O(h^2).
\end{equation}
\label{proposition}
\end{proposition}

\vspace{-4mm}
The detailed proof is provided in Appendix B.
The proposition shows that, unlike traditional RGE, which estimates gradient direction, our sensitivity metric captures the magnitude of local response. Specifically, $\mathbb{E}_u [\|J(x) u\|_2]$ quantifies how much the output changes on average under random perturbations, providing a scalar measure of the token influence.

\begin{algorithm}[t]
\small
\caption{\ours
}
\label{alg:zoo_prune}
\begin{algorithmic}[1]
\State \textbf{Input:} Vision tokens $X \in \mathbb{R}^{N_v \times d_v}$, projection $M$, number of tokens to select $k$, step size $h$, number of perturbations $m$
\State \textbf{Output:} Selected token indices $\mathcal{P}$

\State \textcolor{ForestGreen}{\% ----------- [ZOO-based Sensitivity Estimation] -------------}
    \State Sample $m$ random perturbations $U \in \mathbb{R}^{m \times d_v}$, $u_j \sim \mathcal{N}(0,I)$, normalized to $\|u_j\|_2=1$
    \State Expand $X$ along perturbations: $X^+ = X + hU, \;\; X^- = X - hU$
    \State Project perturbed features: $Z^+ = M(X^+), \;\; Z^- = M(X^-)$
    \State Compute finite-difference responses: $\Delta = \frac{Z^+ - Z^-}{2h}$
    \State Sensitivity: $S(i) = \frac{1}{m}\sum_{j=1}^m \|\Delta_{i,j}\|_2$
\State \textcolor{ForestGreen}{\% ----------- [Sensitivity-Aware Diversity Selection ] ---------}
    \State Normalize sensitivities: 
$\widehat{S}(i) = \frac{S(i) - \min_j S(j)}{\max_j S(j) - \min_j S(j)}$
\State Initialize $\mathcal{P} \gets \emptyset$
\While{$|\mathcal{P}| < k$}
    \State Compute diversity: $\mathrm{Div}(i,\mathcal{P}) = 1 - \max_{j \in \mathcal{P}} \cos(Z_i,Z_j)$ \; (set to $1$ if $\mathcal{P}$ is empty)
    \State Fusion score: $\mathrm{Score}(i) = \widehat{S}(i) \cdot \mathrm{Div}(i,\mathcal{P})$
    \State Select $i^\star = \arg\max_i \mathrm{Score}(i)$
    \State $\mathcal{P} \gets \mathcal{P} \cup \{i^\star\}$
\EndWhile

\State \Return $\mathcal{P}$
\end{algorithmic}
\end{algorithm}

\paragraph{Sensitivity-aware diversity selection.} 
While sensitivity captures the most informative tokens, it does not by itself enforce coverage over diverse visual content.  
To reduce redundancy, we integrate a diversity criterion inspired by DivPrune~\citep{Divprune}.  
Let $Z_i$ denote the vision feature of token $i$, and $\mathcal{P}$ the set of already selected tokens. We define
\begin{equation}
    \mathrm{Div}(i,\mathcal{P}) = 1 - \max_{j \in \mathcal{P}} \cos(Z_i,Z_j),
\end{equation}
where $\cos(\cdot,\cdot)$ is cosine similarity.  
The final selection score is defined as
\begin{equation}
    \mathrm{Score}(i) = \widehat{\mathrm{S}}(i) \cdot \mathrm{Div}(i,\mathcal{P}),
\end{equation}
where $\widehat{\mathrm{S}}(i)$ is the normalized sensitivity score. 
The multiplicative design avoids introducing additional hyperparameters for weighting the two criteria.  
Compared to DivPrune, our diversity selection method introduces two key modifications: 
(1) For the first token, DivPrune selects the one that is maximally distant from all others, whereas ours prioritizes the token with the highest sensitivity.  
(2) For {subsequent} selections, DivPrune considers only diversity, while our method combines sensitivity and diversity via the hybrid score above.  This procedure yields a token subset that is both sensitivity-driven and diversity-driven.
The overall process is described in  Algorithm~\ref{alg:zoo_prune}.

\section{Experiments}
We evaluate \ours on LLaVA-v1.5-7B/13B~\citep{LLaVA-1.5}, LLaVA-NeXT-7B~\citep{LLaVA-NeXT}, and Qwen2.5-VL-7B~\citep{Qwen2.5}. LLaVA-v1.5 employs a CLIP ViT-L~\citep{CLIP} with 576 tokens, LLaVA-NeXT scales to 2880 tokens for high-resolution inputs, and Qwen2.5-VL adopts a dynamic-resolution ViT encoder. Following VisionZip~\citep{visionzip}, we evaluate multiple pruning ratios and report performance relative to the unpruned baseline across nine benchmarks. For LLaVA-NeXT, we applied a low-rank factorization ($k=128$) to the MM-projector layers to further boost efficiency, since this model processes a large number of visual tokens.
All experiments are training-free and calibration-free, run on 4$\times$A6000 GPUs with $m{=}64$, $h{=}0.01$, and evaluated using \texttt{lmms-eval}~\citep{Lmms-eval}.
More details in Appendix C.

\begin{table*}[t!]
\scriptsize

\centering
\caption{Performance Comparison on LLaVA-1.5-7B.}
\vspace{-3mm}
\label{tab:LLaVA-1.5-7B_more}
\resizebox{1.0\textwidth}{!}{
\setlength{\tabcolsep}{7pt}
\begin{tabular}{l|ccccccc|c}
\toprule
\textbf{Method} & 
\begin{tabular}[c]{@{}c@{}} \textbf{GQA} \\ Acc. $\uparrow$ \end{tabular} & 
\begin{tabular}[c]{@{}c@{}} \textbf{MMB} \\ Acc. $\uparrow$ \end{tabular} & 
\begin{tabular}[c]{@{}c@{}} \textbf{MME} \\ P+C $\uparrow$ \end{tabular} & 
\begin{tabular}[c]{@{}c@{}} \textbf{POPE} \\ F1 $\uparrow$ \end{tabular} & 
\begin{tabular}[c]{@{}c@{}} \textbf{SQA} \\ Acc. $\uparrow$ \end{tabular} & 
\begin{tabular}[c]{@{}c@{}} \textbf{VQA$^{V2}$} \\ Acc. $\uparrow$ \end{tabular} & 
\begin{tabular}[c]{@{}c@{}} \textbf{VQA$^{Text}$} \\ Acc. $\uparrow$ \end{tabular} &
\begin{tabular}[c]{@{}c@{}} \textbf{Avg.} \\ $\uparrow$ \end{tabular} \\
\midrule
\rowcolor{gray!20}
\multicolumn{9}{c}{\textit{Total 576 Tokens}} \\
\midrule

LLaVA-1.5-7B & 61.90 & 64.70 & 1862.00 & 85.90 & 69.50 & 78.50 & 58.20 & 100\% \\
\midrule
\rowcolor{gray!20}
\multicolumn{9}{c}{\textit{Retain 192 Tokens} \textcolor{ForestGreen}{$\downarrow$ 66.7\%}} \\
\midrule
    ToMe \scriptsize{\textcolor{gray}{(ICLR 2023)}} & 54.30 & 60.50 & 1563.00 & 72.40 & 65.20 & 68.00 & 52.10 & 88.49\% \\
    FastV \scriptsize{\textcolor{gray}{(ECCV 2024)}} & 52.70 & 61.20 & 1612.00 & 64.80 & 67.30 & 67.10 & 52.50 & 87.75\% \\
    HiRED \scriptsize{\textcolor{gray}{(AAAI 2025)}} & 58.70 & 62.80 & 1737.00 & 82.80 & 68.40 & 74.90 & 47.40 & 93.84\% \\
    VisionZip \scriptsize{\textcolor{gray}{(CVPR 2025)}} & 59.30 & 63.00 & \textbf{1782.60} & 85.30 & 68.90 & 76.80 & 57.30 & 97.66\% \\
    DivPrune \scriptsize{\textcolor{gray}{(CVPR 2025)}} & 59.97 & 62.54 & 1762.23 & 87.00 & 68.91 & 76.87 & 56.97 & 97.78\% \\
    PyramidDrop \scriptsize{\textcolor{gray}{(CVPR 2025)}} & 57.10 & \textbf{63.20} & 1766.00 & 82.30 & 68.80 & 75.10 & 56.10 & 95.95\% \\
    SparseVLM \scriptsize{\textcolor{gray}{(ICML 2025)}} & 57.60 & 62.50 & 1721.00 & 83.60 & 69.10 & 75.60 & 56.10 & 95.93\% \\
    \textbf{\ours (Ours)} & \textbf{60.03} & 62.89 & 1781.66 & \textbf{87.24} & \textbf{69.16} & \textbf{77.34} & \textbf{57.30} & \textcolor[HTML]{EB4949}{\textbf{98.27}\%} \\
\midrule
\rowcolor{gray!20}
\multicolumn{9}{c}{\textit{Retain 128 Tokens} \textcolor{ForestGreen}{$\downarrow$ 77.8\%}} \\
\midrule
    ToMe \scriptsize{\textcolor{gray}{(ICLR 2023)}} & 52.40 & 53.30 & 1343.00 & 62.80 & 59.60 & 63.00 & 49.10 & 80.38\% \\
    FastV \scriptsize{\textcolor{gray}{(ECCV 2024)}} & 49.60 & 56.10 & 1490.00 & 59.60 & 60.20 & 61.80 & 50.60 & 81.22\% \\
    HiRED \scriptsize{\textcolor{gray}{(AAAI 2025)}} & 57.20 & 61.50 & 1710.00 & 79.80 & 68.10 & 73.40 & 46.10 & 91.84\% \\
    VisionZip \scriptsize{\textcolor{gray}{(CVPR 2025)}} & 57.60 & 62.00 & \textbf{1761.70} & 83.20 & 68.90 & 75.60 & 56.80 & 96.20\% \\
    DivPrune \scriptsize{\textcolor{gray}{(CVPR 2025)}} & 59.25 & \textbf{62.03} & 1718.22 & 86.72 & \textbf{68.96} & 75.96 & 56.06 & 96.73\% \\
    PyramidDrop \scriptsize{\textcolor{gray}{(CVPR 2025)}} & 56.00 & 61.00 & 1644.00 & 82.30 & 68.30 & 72.90 & 55.10 & 93.52\% \\
    SparseVLM \scriptsize{\textcolor{gray}{(ICML 2025)}} & 56.00 & 60.00 & 1696.00 & 80.50 & 67.10 & 73.80 & 54.90 & 93.27\% \\
    \textbf{\ours (Ours)} & \textbf{59.49} & 61.86 & 1751.60 & \textbf{87.13} & 68.91 & \textbf{76.57} & \textbf{57.87} & \textcolor[HTML]{EB4949}{\textbf{97.62}\%} \\

\midrule
\rowcolor{gray!20}
\multicolumn{9}{c}{\textit{Retain 64 Tokens} \textcolor{ForestGreen}{$\downarrow$ 88.9\%}} \\
\midrule
    ToMe \scriptsize{\textcolor{gray}{(ICLR 2023)}} & 48.60 & 43.70 & 1138.00 & 52.50 & 50.00 & 57.10 & 45.30 & 70.12\% \\
    FastV \scriptsize{\textcolor{gray}{(ECCV 2024)}} & 46.10 & 48.00 & 1256.00 & 48.00 & 51.10 & 55.00 & 47.80 & 71.10\% \\
    HiRED \scriptsize{\textcolor{gray}{(AAAI 2025)}} & 54.60 & 60.20 & 1599.00 & 73.60 & 68.20 & 69.70 & 44.20 & 87.95\% \\
    VisionZip \scriptsize{\textcolor{gray}{(CVPR 2025)}} & 55.10 & 60.10 & \textbf{1690.00} & 77.00 & \textbf{69.00} & 72.40 & \textbf{55.50} & 92.74\% \\
    DivPrune \scriptsize{\textcolor{gray}{(CVPR 2025)}} & 57.78 & 59.28 & 1674.40 & 85.56 & 68.17 & 74.11 & 54.69 & 94.42\% \\
    PyramidDrop \scriptsize{\textcolor{gray}{(CVPR 2025)}} & 41.90 & 33.30 & 1092.00 & 55.90 & 68.60 & 69.20 & 45.90 & 72.66\% \\
    SparseVLM \scriptsize{\textcolor{gray}{(ICML 2025)}} & 52.70 & 56.20 & 1505.00 & 75.10 & 62.20 & 68.20 & 51.80 & 86.52\% \\
    \textbf{\ours (Ours)} & \textbf{58.47} & \textbf{60.22} & 1675.59 & \textbf{85.86} & 68.27 & \textbf{75.02} & 55.35 & \textcolor[HTML]{EB4949}{\textbf{95.20}\%} \\

\bottomrule
\bottomrule
\end{tabular}
\vspace{-5mm}
}
\end{table*}
\subsection{Comparison on Diverse Tasks}
\paragraph{Results on  LLaVA-1.5-7B.}
As shown in Table~\ref{tab:LLaVA-1.5-7B_more}, \ours consistently outperforms the state-of-the-art training-free pruning methods, especially under aggressive compression. On LLaVA-1.5-7B, it preserves 95.20\% performance with only 64 tokens, surpassing DivPrune~\citep{Divprune} (94.40\%) and far exceeding attention-based methods such as FastV~\citep{FastV}, which retains three times more tokens but drops to 87.75\%. Additional results on LLaVA-1.5-13B in Appendix D.

\begin{figure*}[t]
    \begin{minipage}[h]{0.47\linewidth}
        \centering
         \vspace*{-1mm}
    \captionof{table}{Performance comparison on Qwen2.5-VL-7B.}
        \label{tab:qwen25vl}
        \vspace{-3mm}
        \resizebox{\linewidth}{!}{
        \setlength{\tabcolsep}{4pt}
        \begin{tabular}{l|cccc|c}
        \toprule
        \textbf{Method} & GQA & MMB & MME & POPE& Avg. \\
        \midrule
        \rowcolor{gray!20}
        \multicolumn{6}{c}{\textit{Baseline (Full Tokens)}} \\
        \midrule
        Qwen2.5-VL-7B & 60.84 & 84.10 & 2310 & 86.3 & 100\% \\
        \midrule
        \rowcolor{gray!20}
        \multicolumn{6}{c}{\textit{Retain 20\% Tokens}} \\
        \midrule
        VisionZip {\small \textcolor{gray}{(CVPR 2025)}}& 57.27	&79.72&	\textbf{2221}	&83.89 &  95.6\% \\
        DivPrune {\small\textcolor{gray}{(CVPR 2025)}}& \textbf{60.05}&	79.55&	2173	&83.42&  96.0\% \\
        \textbf{\ours (Ours)} & {58.81} & \textbf{80.60} & {2201} & \textbf{84.17}  & \textcolor[HTML]{EB4949}{\textbf{96.2}}\% \\
        \midrule
        \rowcolor{gray!20}
        \multicolumn{6}{c}{\textit{Retain 10\% Tokens}} \\
        \midrule
        VisionZip {\small\textcolor{gray}{(CVPR 2025)}} & 54.09&	76.03&	1937	&78.97 & 88.7\% \\
        DivPrune {\small\textcolor{gray}{(CVPR 2025)}} & \textbf{55.49}&	76.03	&\textbf{2054}	&79.05 & 90.5\% \\
        \textbf{\ours (Ours)} & {55.45} & \textbf{76.28} & {2018} & \textbf{80.99}  & \textcolor[HTML]{EB4949}{\textbf{90.8}}\% \\
        \bottomrule
        \end{tabular}
        }
    \end{minipage}
         \hspace{0.038\linewidth}
    \begin{minipage}[h]{0.49\linewidth}
             \vspace*{-1mm}
        \centering
        \setlength{\abovecaptionskip}{5.5pt}
        \includegraphics[width=0.93\linewidth]{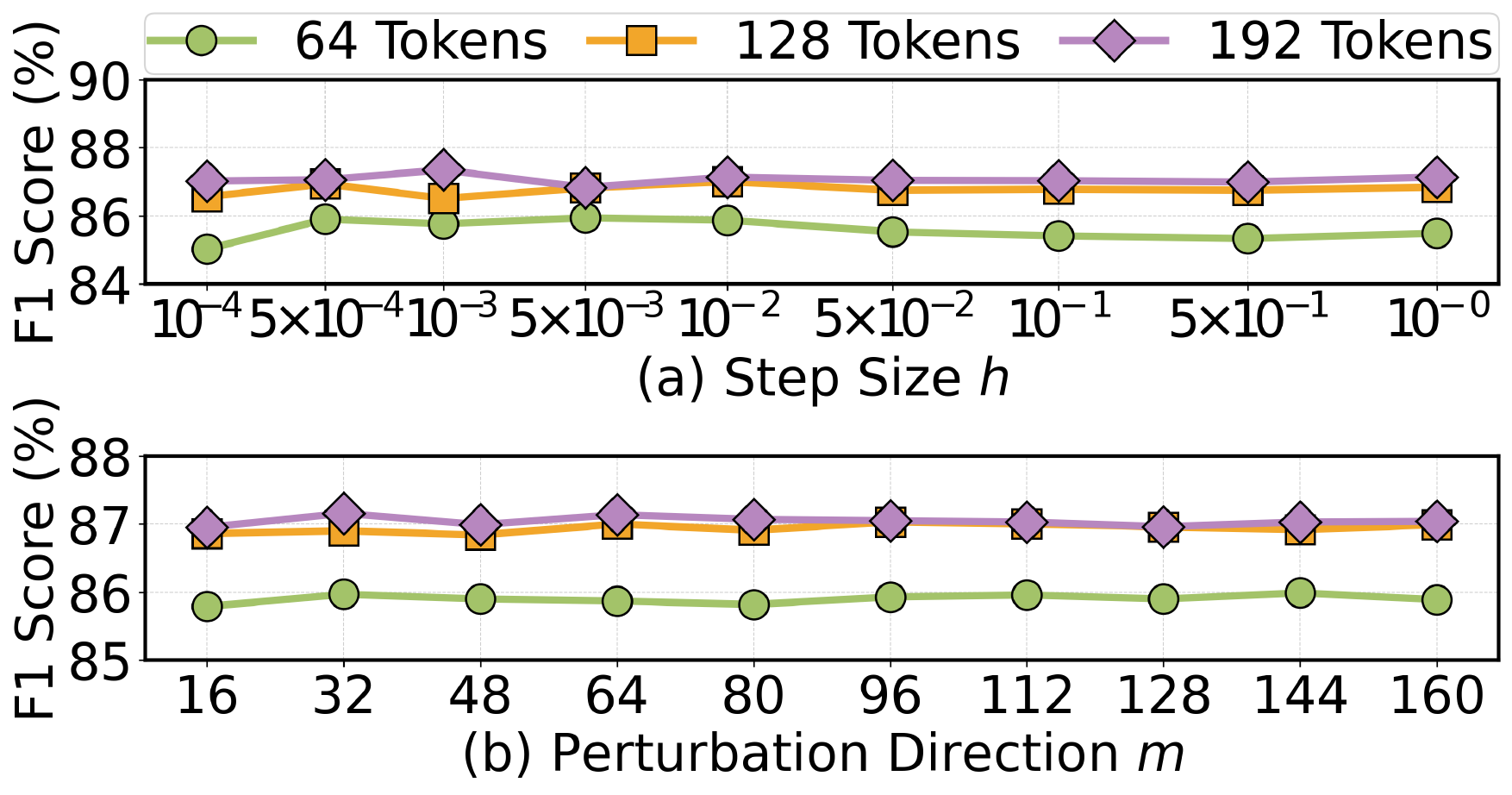}
        \caption{
        Hyperparameter sensitivity on POPE with LLaVA-1.5-7B: (a) small step size $h$, (b) number of perturbation directions $m$.
        }
        \label{fig:hyperparam}
    \end{minipage}
    \vspace{-4mm}
\end{figure*}

\vspace{-4mm}
\begin{table*}[t!]
\footnotesize
\vspace{-3mm}
\centering
\caption{Performance Comparison on LLaVA-NeXT-7B.}
\vspace{-3mm}
\label{tab:LLaVA-NeXT-7B}
\resizebox{\textwidth}{!}{
\setlength{\tabcolsep}{6 pt}
\begin{tabular}{l|ccccccccc|c} 
\toprule
\textbf{Method} & 
\begin{tabular}[c]{@{}c@{}} \textbf{GQA} \\ Acc. $\uparrow$ \end{tabular} & 
\begin{tabular}[c]{@{}c@{}} \textbf{MMB} \\ Acc. $\uparrow$ \end{tabular} & 
\begin{tabular}[c]{@{}c@{}} \textbf{MME} \\ P+C $\uparrow$ \end{tabular} & 
\begin{tabular}[c]{@{}c@{}} \textbf{POPE} \\ F1 $\uparrow$ \end{tabular} & 
\begin{tabular}[c]{@{}c@{}} \textbf{SQA} \\ Acc. $\uparrow$ \end{tabular} & 
\begin{tabular}[c]{@{}c@{}} \textbf{VQA$^{V2}$} \\ Acc. $\uparrow$ \end{tabular} & 
\begin{tabular}[c]{@{}c@{}} \textbf{VQA$^{Text}$} \\ Acc. $\uparrow$ \end{tabular} & 
\begin{tabular}[c]{@{}c@{}} \textbf{MMMU} \\ Acc. $\uparrow$ \end{tabular} & 
\begin{tabular}[c]{@{}c@{}} \textbf{SEED-I} \\ Acc. $\uparrow$ \end{tabular} & 
\begin{tabular}[c]{@{}c@{}} \textbf{Avg.} \\ $\uparrow$ \end{tabular} \\
\midrule
\rowcolor{gray!20}
\multicolumn{11}{c}{\textit{Total 2880 Tokens}} \\
\midrule

LLaVA-NeXT-7B & 64.20 & 67.90 & 1842.00 & 86.40 & 70.20 & 80.10 & 61.30 & 35.10 & 70.20 & 100\% \\
 
\midrule
\rowcolor{gray!20}
\multicolumn{11}{c}{\textit{Retain 640 Tokens} \textcolor{mygreen}{$\downarrow$ 77.8\%}} \\
\midrule
SparseVLM {\scriptsize  \textcolor{gray}{(ICML 2025)}} & 60.30 & 65.70 & 1772.00 & -- & 67.70 & 77.10 & 57.80 & 34.60 & -- & -- \\
VisionZip {\scriptsize  \textcolor{gray}{(CVPR 2025)}} & 61.30 & \textbf{66.30} & 1787.00 & 86.30 & \textbf{68.10} & 79.10 & \textbf{60.20} & 34.70 & 66.70 & 97.5\% \\
DivPrune {\scriptsize  \textcolor{gray}{(CVPR 2025)}} & 61.58 & 65.38 & 1773.04 & 85.51 & 67.82 & 78.94 & 55.41 & 36.89 & 67.56 & 97.1\% \\
\textbf{\ours (Ours)} & \textbf{62.19} & 65.21 & \textbf{1816.45} & \textbf{86.75} & 68.02	& \textbf{79.64} & 57.98 & \textbf{36.89} & \textbf{67.95} & \textcolor[HTML]{EB4949}{\textbf{98.3}}\% \\

\midrule
\rowcolor{gray!20}
\multicolumn{11}{c}{\textit{Retain 320 Tokens} \textcolor{mygreen}{$\downarrow$ 88.9\%}} \\
\midrule
SparseVLM {\scriptsize  \textcolor{gray}{(ICML 2025)}} & 57.70 & 64.30 & 1694.00 & -- & 67.30 & 73.40 & 55.90 & 34.40 & -- & -- \\
VisionZip {\scriptsize  \textcolor{gray}{(CVPR 2025)}} & 59.30 & 63.10 & 1702.00 & 82.10 & 67.30 & 76.20 & \textbf{58.90} & 35.30 & 63.40 & 94.5\% \\
DivPrune {\scriptsize  \textcolor{gray}{(CVPR 2025)}} & 59.63 & 63.66 & 1731.04 & 83.47 & \textbf{67.82} & 76.64 & 53.84 & \textbf{37.11} & 65.35 & 95.1\% \\
\textbf{\ours (Ours)} & \textbf{60.97} & \textbf{64.86} & \textbf{1787.68}	& \textbf{85.47} & 67.77 & \textbf{78.08} & 57.28 & 37.00 & \textbf{66.47} & \textcolor[HTML]{EB4949}{\textbf{97.1}}\%\\

\midrule
\rowcolor{gray!20}
\multicolumn{11}{c}{\textit{Retain 160 Tokens} \textcolor{mygreen}{$\downarrow$ 94.4\%}} \\
\midrule
SparseVLM {\scriptsize  \textcolor{gray}{(ICML 2025)}} & 51.20 & 63.10 & 1542.00 & -- & 67.50 & 66.30 & 46.40 & 32.80 & -- & -- \\
VisionZip {\scriptsize  \textcolor{gray}{(CVPR 2025)}} & 55.50 & 60.10 & 1630.00 & 74.80 & 68.30 & 71.40 & \textbf{56.20} & 36.10 & 58.30 & 90.4\% \\
DivPrune {\scriptsize  \textcolor{gray}{(CVPR 2025)}} & 57.79 & 62.29 & 1658.25 & 79.36 & 68.02 & 73.92 & 52.42 & 36.44 & 62.54 & 92.4\% \\
\textbf{\ours (Ours)} & \textbf{59.93} & \textbf{64.18} & \textbf{1738.64} & \textbf{83.05}	& \textbf{68.42} & \textbf{76.12} & 55.42 & \textbf{37.11} & \textbf{64.05} & \textcolor[HTML]{EB4949}{\textbf{95.4}}\%\\

\bottomrule
\bottomrule
\end{tabular}
\vspace{-4mm}
}
\end{table*}

\paragraph{Results on LLaVA-NeXT-7B.}
We further evaluate \ours on LLaVA-NeXT-7B, which processes up to 2880 tokens. As reported in Table~\ref{tab:LLaVA-NeXT-7B}, \ours maintains 98.3\% performance when pruning 77.8\% of tokens, outperforming VisionZip and nearly matching the baseline. Even with a 94.4\% pruning (160 tokens), it can maintain a 95.4\% performance, demonstrating the scalability of sensitivity-aware diversity selection under high-resolution compression.

\vspace{-4mm}
\paragraph{Results on Qwen2.5-VL-7B.}
To validate the generalizability of \ours, we tested it on Qwen2.5-VL-7B, a distinct VLM variant beyond LLaVA family with dynamic-resolution inputs. As shown in Table~\ref{tab:qwen25vl}, \ours continues to lead, achieving 96.2\% performance with a 20\% token budget and 90.8\% with only 10\% of the tokens, consistently surpassing both VisionZip and DivPrune. These results confirm that \ours\ generalizes robustly across diverse VLM architectures, effectively preserving task performance even with long, variable-length token sequences.

\subsection{Ablation Studies and Further Analysis}

\begin{table*}[t!]
\footnotesize
\vspace{-1mm}
\centering
\caption{Ablation on Token Selection Metrics with LLaVA-NeXT-7B.}
\vspace{-3mm}
\label{tab:LLaVA-next_ablation}
\resizebox{\textwidth}{!}{
\setlength{\tabcolsep}{7 pt}
\begin{tabular}{ccc|ccccccccc|c} 
\toprule
\textbf{Sensitivity} & \textbf{Diversity} & \textbf{Fusion} &
\begin{tabular}[c]{@{}c@{}} \textbf{GQA} \\ Acc. $\uparrow$ \end{tabular} & 
\begin{tabular}[c]{@{}c@{}} \textbf{MMB} \\ Acc. $\uparrow$ \end{tabular} & 
\begin{tabular}[c]{@{}c@{}} \textbf{MME} \\ P+C $\uparrow$ \end{tabular} & 
\begin{tabular}[c]{@{}c@{}} \textbf{POPE} \\ F1 $\uparrow$ \end{tabular} & 
\begin{tabular}[c]{@{}c@{}} \textbf{SQA} \\ Acc. $\uparrow$ \end{tabular} & 
\begin{tabular}[c]{@{}c@{}} \textbf{VQA$^{V2}$} \\ Acc. $\uparrow$ \end{tabular} & 
\begin{tabular}[c]{@{}c@{}} \textbf{VQA$^{Text}$} \\ Acc. $\uparrow$ \end{tabular} & 
\begin{tabular}[c]{@{}c@{}} \textbf{MMMU} \\ Acc. $\uparrow$ \end{tabular} & 
\begin{tabular}[c]{@{}c@{}} \textbf{SEED-I} \\ Acc. $\uparrow$ \end{tabular} & 
\begin{tabular}[c]{@{}c@{}} \textbf{Avg.} \\ $\uparrow$ \end{tabular} \\

\midrule
\rowcolor{gray!20}
\multicolumn{13}{c}{\textit{Retain 640 Tokens} \textcolor{ForestGreen}{$\downarrow$ 77.8\%}} \\
\midrule

$\checkmark$ &  & - & 61.23 & 65.21	& 1818.62 & 86.54 & 68.07 & 78.47 & 54.12 & 35.78 & 66.29 & 96.7\% \\
& $\checkmark$ & - & 61.58 & 65.38 & 1773.04 & 85.51 & 67.82 & 78.94 & 55.41 & 36.89 & 67.56 & 97.1\% \\
$\checkmark$ & $\checkmark$ & Sum & 61.81 & \textbf{65.55} & 1794.17 & 86.34	& \textbf{68.27} & 79.38 & \textbf{57.99} & \textbf{37.22}	& 67.29	& 98.1\%\\
$\checkmark$ & $\checkmark$ & Multiply & \textbf{62.19} & 65.21 & \textbf{1816.45} & \textbf{86.75} & 68.02 & \textbf{79.64} & 57.98 & 36.89 & \textbf{67.95} & \textcolor[HTML]{EB4949}{\textbf{98.3}}\% \\

\midrule
\rowcolor{gray!20}
\multicolumn{13}{c}{\textit{Retain 320 Tokens} \textcolor{ForestGreen}{$\downarrow$ 88.9\%}} \\
\midrule
$\checkmark$ & & - & 59.22 & 64.69 & 1744.42 & 83.15 & 67.63 & 75.69 & 47.25 & 34.78 & 63.69 & 92.9\% \\
& $\checkmark$ & - & 59.63 & 63.66 & 1731.04 & 83.47 & 67.82 & 76.64 & 53.84 & 37.11 & 65.35 & 95.1\% \\
$\checkmark$ & $\checkmark$ & Sum & 60.47 & 64.43 & 1761.52	& 84.21	& \textbf{68.47} & 77.70	& 56.79	& 36.89	& 65.46	& 96.4\% \\
$\checkmark$ & $\checkmark$ & Multiply & \textbf{60.97} & \textbf{64.86} & \textbf{1787.68}	& \textbf{85.47} & 67.77 & \textbf{78.08} & \textbf{57.28} & \textbf{37.00} & \textbf{66.47} & \textcolor[HTML]{EB4949}{\textbf{97.1}}\%\\

\midrule
\rowcolor{gray!20}
\multicolumn{13}{c}{\textit{Retain 160 Tokens}\textcolor{ForestGreen}{$\downarrow$ 94.4\%}} \\
\midrule
$\checkmark$ & & - & 57.23 & 61.86 &  1674.35 & 77.27 & 68.82 & 72.58 & 50.96 & 35.56 & 61.04 & 91.2\%\\
& $\checkmark$ & - & 57.79 & 62.29 & 1658.25 & 79.36 & 68.02 & 73.92 & 52.42 & 36.44 & 62.54 & 92.4\% \\
$\checkmark$ & $\checkmark$ & Sum & 59.48 & 63.49 & 1710.35	& 81.21	& 68.22	& 75.88	& 54.77	& 34.78	& 63.47	& 93.8\% \\
$\checkmark$ & $\checkmark$ & Multiply & \textbf{59.93} & \textbf{64.18} & \textbf{1738.64} & \textbf{83.05} & \textbf{68.42} & \textbf{76.12} & \textbf{55.42} & \textbf{37.11} & \textbf{64.05} & \textcolor[HTML]{EB4949}{\textbf{95.4}}\%\\

\bottomrule
\bottomrule
\end{tabular}
\vspace{-9mm}
}
\end{table*}

\paragraph{Ablation Studies.}
We ablate four variants on LLaVA-NeXT-7B (Table~\ref{tab:LLaVA-next_ablation}): Sensitivity-only, Diversity-only (DivPrune), and two Fusion strategies (sum, multiply). Sensitivity-only selects task-critical tokens and excels on reasoning, but lags on TextVQA where context is vital. Diversity-only offers broad coverage yet misses key cues. Combining both is consistently better: Fusion (Multiply) delivers 98.3\% at 22.2\% tokens without extra hyperparameters.
It means that sensitivity and diversity are complementary and jointly essential for training-free token pruning.

\begin{figure*}[!t]
    \vspace{-0.3cm}
    \centering
    \setlength{\abovecaptionskip}{0.05cm}
    \includegraphics[width=\linewidth]{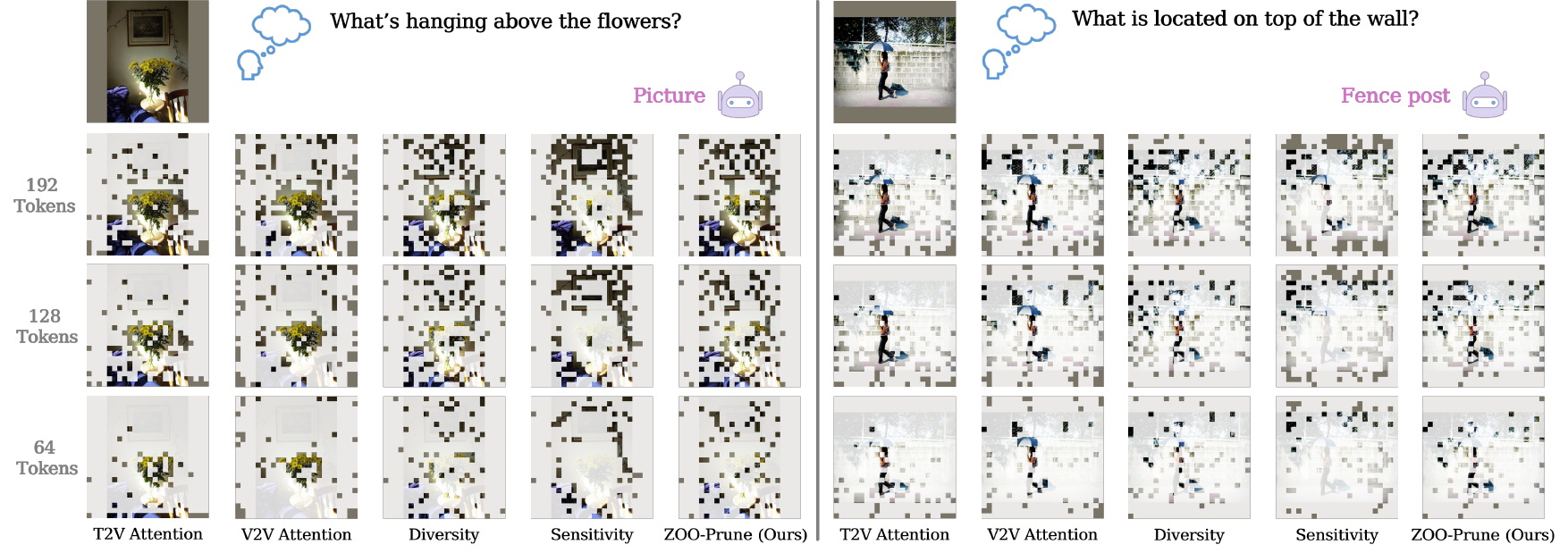}
\caption{
Qualitative comparison on the GQA benchmark. Attention-based methods suffer from positional bias or redundant token clusters. Diversity-based pruning spreads tokens broadly but lacks semantic focus. The ZOO-based sensitivity captures output-related tokens but overlooks spatial coverage. Our ZOO-Prune jointly optimizes sensitivity and diversity for balanced selection across compression ratios.
}
    \label{fig:exp:visualization_2}
\end{figure*}

\begin{figure*}[t]
    \begin{minipage}[h]{0.47\linewidth}
        \centering
         \vspace*{-1mm}
         \setlength{\abovecaptionskip}{5.5pt}
            \includegraphics[width=0.92\linewidth]{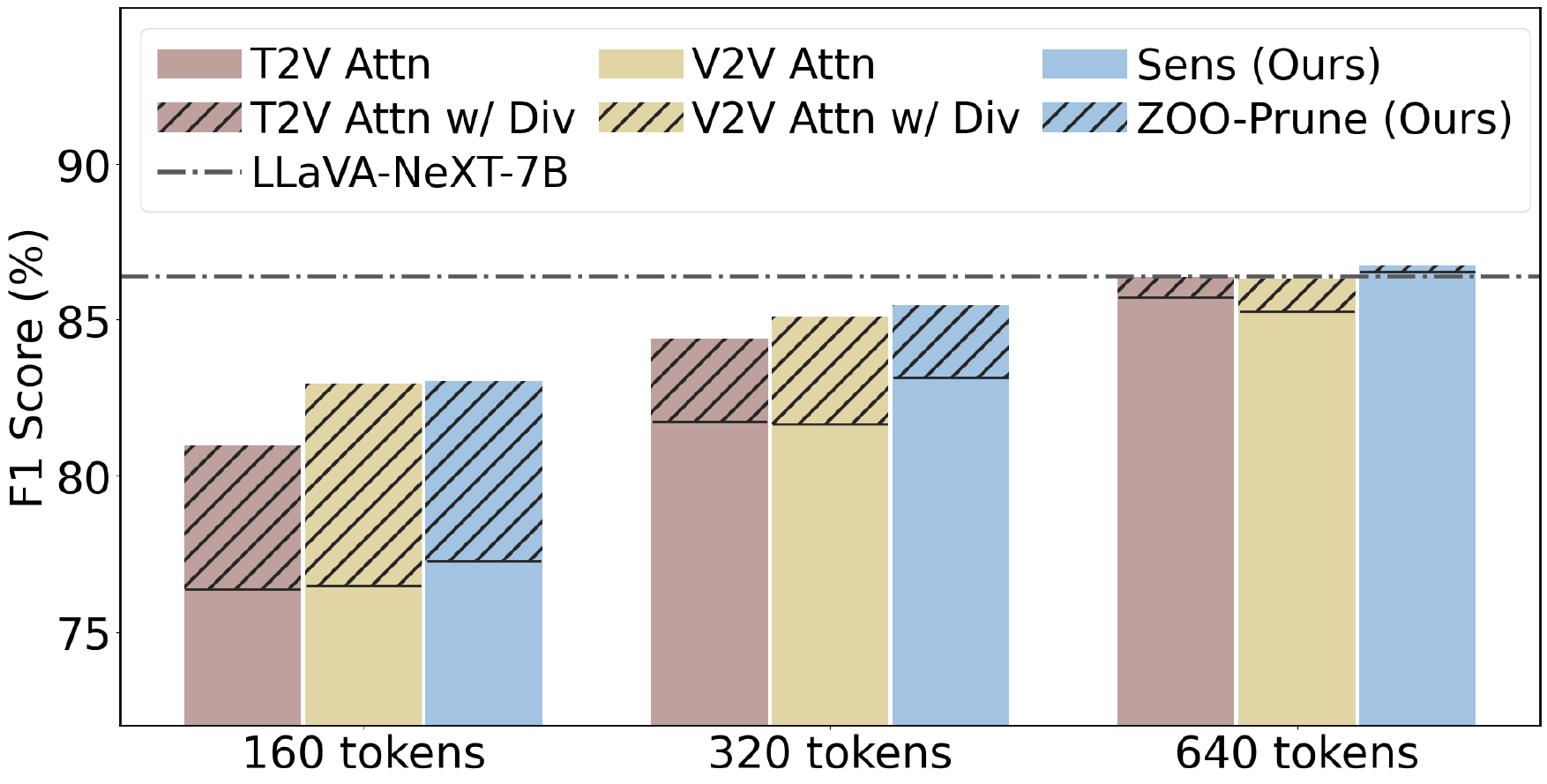}
            \caption{
           ZOO Sensitivity vs. Attention Pruning.  ZOO-based sensitivity (Sens) metric consistently outperforms both text-visual (T2V Attn) and visual-visual (V2V Attn) attention-based pruning, even when combined with diversity (Div).
            }
            \label{fig:sens_vs_attn_comparison}
    \end{minipage}
         \hspace{0.008\linewidth}
    \begin{minipage}[h]{0.53\linewidth}
             \vspace*{-1mm}
                \begin{subfigure}[b]{0.5\columnwidth}
                        \includegraphics[width=\textwidth]{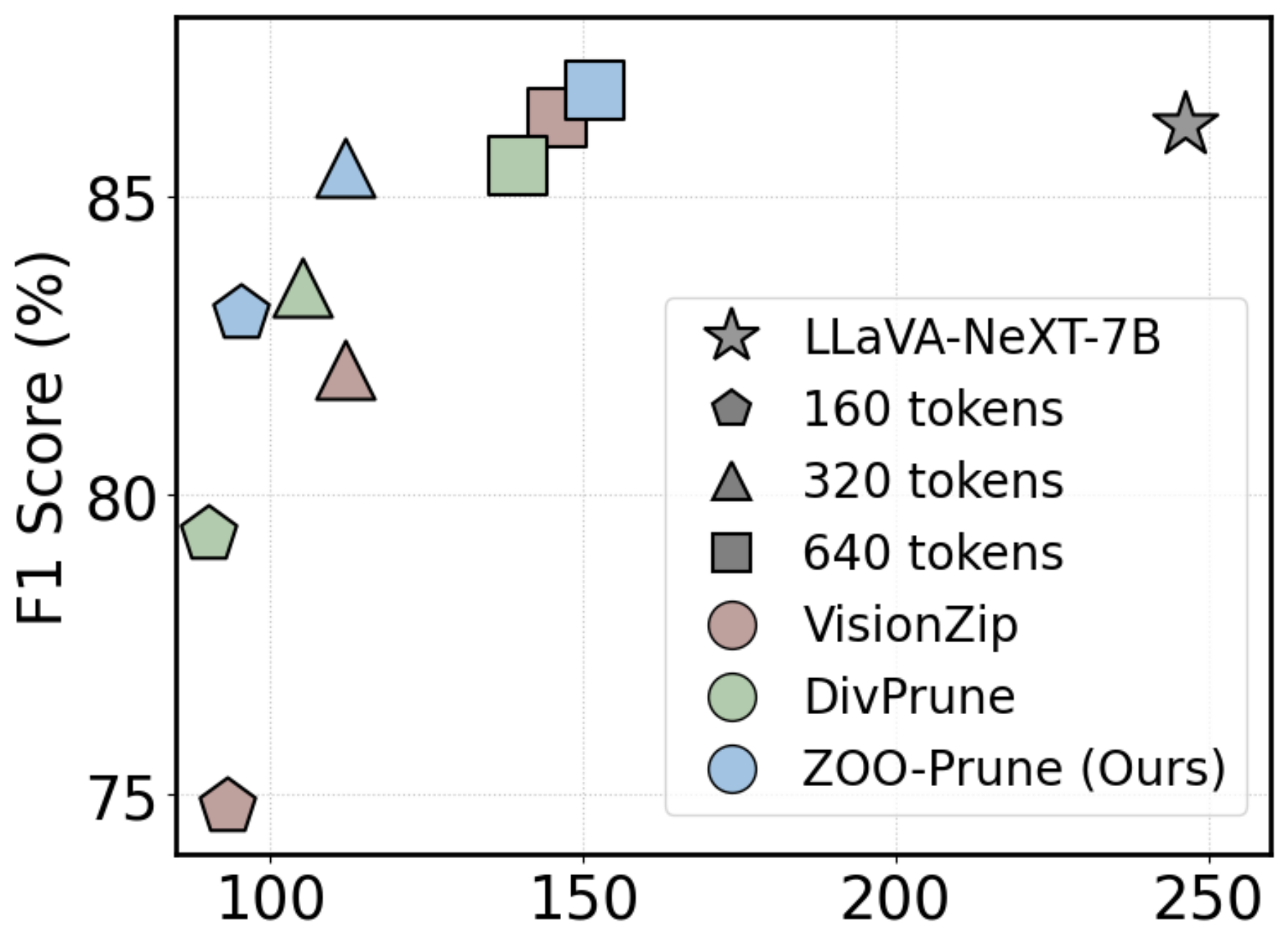}
                        \caption{Prefilling Time (ms)}
                        \label{fig:prefilling}
                    \end{subfigure}
                    \begin{subfigure}[b]{0.5\columnwidth}
                        \includegraphics[width=\textwidth]{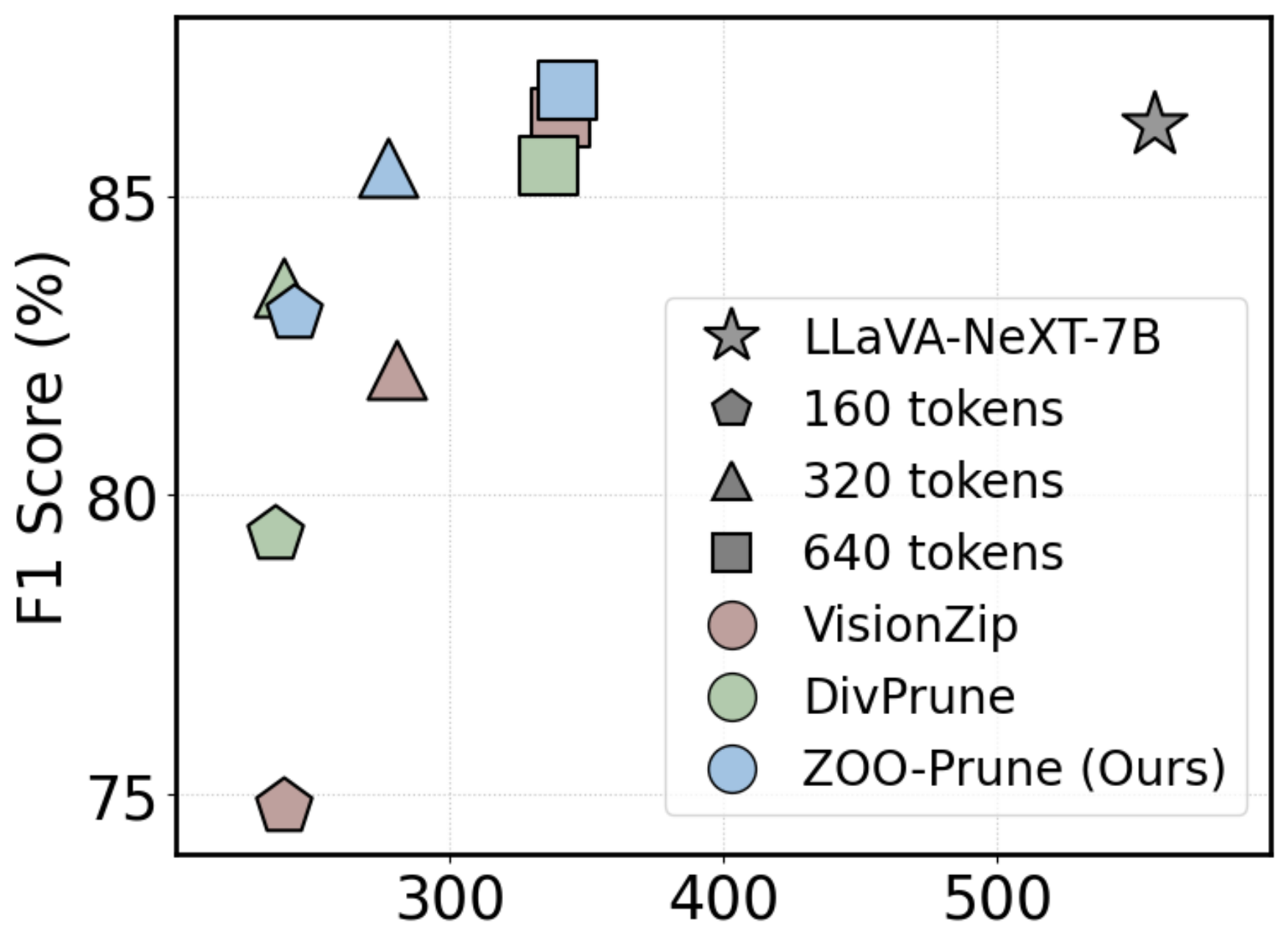}
                        \caption{E2E Latency (secs)}
                        \label{fig:latency}
                    \end{subfigure}
                    \vspace{-1mm}
                    \setlength{\abovecaptionskip}{-3pt}
                    \caption{Inference efficiency on the POPE benchmark relative to the LLaVA-NeXT-7B baseline. The left scatter plot reports prefilling time, and the right scatter plot shows end-to-end (E2E) latency. Each point represents a method–token-count pair, illustrating the trade-off between computation cost and F1 score performance.}
                    \label{fig:performance_comparison}
    \end{minipage}
\end{figure*}


\paragraph{Hyperparameter Analysis.}


We analyze the sensitivity of \ours\ to its two hyperparameters $m$ (number of perturbations) and $h$ (step size) on the POPE benchmark (Fig.~\ref{fig:hyperparam}). Performance is stable across a wide range: accuracy remains consistent for $m=16$–$160$ and is similarly insensitive to $h$ over $1\text{e-}4$–$1$. We adopt $m=64$ to reduce variance at negligible cost, and fix $h=0.01$ for all experiments, eliminating the need for task-specific tuning.

\vspace{-4mm}
\paragraph{Revisiting Token Importance: Sensitivity vs. Attention vs. Diversity.}
We compare our zeroth-order sensitivity metric against two main categories of training-free baselines: attention-based and diversity-based. For the attention-based methods, we implement two variants: (1) a text-visual (T2V) approach using attention scores after layer 3, following FastV~\citep{FastV}, and (2) a visual-visual (V2V) approach like VisionZip~\citep{visionzip} that averages token attention scores. For the diversity-based pruning, we implement the max-min feature diversity selection from DivPrune~\citep{Divprune}.

As shown in Fig.~\ref{fig:exp:visualization_2}, attention-guided pruning suffers from inherently limitations:
T2V attention exhibits positional bias by overweighting tokens physically near the query (often in lower image regions), while V2V attention preserves numerous duplicate tokens.
These issues make attention-based pruning unstable under aggressive compression. Diversity-only pruning improves spatial coverage but treats tokens uniformly, often discarding semantically important regions.
Our ZOO-based sensitivity quantifies each token’s impact on the model output, providing a stable and architecture-agnostic importance signal. 

We also evaluate the behavioral and consistency aspects of each method through qualitative analyses on POPE.
As shown in Fig.~\ref{fig:sens_vs_attn_comparison}, sensitivity-based pruning consistently outperforms attention-driven variants across all compression ratios. When combined with diversity in \ours, it retains high-sensitivity tokens while ensuring sufficient content coverage, achieving a markedly superior accuracy–efficiency trade-off.
Overall, these quantitative and qualitative results show that \ours unifies sensitivity and diversity into a robust and broadly applicable token pruning framework. See more in Appendix~E.


\vspace{-4mm}
\paragraph{Inference Efficiency.}
To assess efficiency, we measure end-to-end (E2E) latency and prefilling time in Fig.~\ref{fig:performance_comparison}. Prefilling, dominated by visual token processing, is the main stage accelerated by pruning. While all methods achieve substantial speedups, \ours consistently demonstrates a superior accuracy–efficiency trade-off: it achieves competitive or better POPE performance while operating at significantly lower latency. At the most aggressive setting (160 tokens), \ours reduces E2E latency by 2.30$\times$ and prefilling time by 2.59$\times$. Importantly, \ours incurs only negligible overhead from sensitivity estimation, yet sustains markedly higher accuracy than competing baselines, making it a practical and effective solution for real-world VLM deployment on resource-constrained devices. All experiments are conducted on a single NVIDIA L40S GPU.
Additional analysis of FLOPs in Appendix D.


\section{Conclusion}
\vspace{-2mm}
In this paper, we presented \ours, a training-free and attention-free token pruning framework for vision–language models that unifies zeroth-order sensitivity estimation with diversity-aware selection. By leveraging projection-layer responses, our method measures token importance without backpropagation, while Sensitivity-Aware Diversity Selection preserves both informativeness and coverage. Extensive experiments across multiple benchmarks show that \ours achieves state-of-the-art accuracy among training-free pruning methods and reduces inference cost with minimal loss. In addition, the method remains stable across architectures and pruning ratios, demonstrating strong generality. Interestingly, even at extremely high pruning levels, the sensitivity-based ranking reliably preserved the same set of task-critical regions across models. We believe these results highlight the practical value of gradient-free sensitivity signals for efficient multimodal learning and point toward broader applications in scalable token and feature compression.


\section*{Acknowledgement.}
This research was supported by the AI Computing Infrastructure Enhancement (GPU Rental Support) User Support Program (RQT-25-110063) funded by the Ministry of Science and ICT (MSIT), Republic of Korea. This research was supported by the MSIT(Ministry of Science, ICT), Korea, under the Global Research Support Program in the Digital Field program(RS-2024-00425354) supervised by the IITP(Institute for Information \& Communications Technology Planning \& Evaluation). This research was supported by the MSIT(Ministry of Science and ICT), Korea, under the Graduate School of Virtual Convergence support program(IITP-2026-RS-2023-00254129) supervised by the IITP(Institute for Information \& Communications Technology Planning \& Evaluation).
{
    \small
    \bibliographystyle{ieeenat_fullname}
    \bibliography{main}
}

\clearpage
\appendix
\renewcommand\thefigure{\Alph{figure}}    
\setcounter{figure}{0}  
\renewcommand\thetable{\Alph{table}}
\setcounter{table}{0} 
\maketitlesupplementary

This appendix supplements the main paper with the theoretical analysis of Proposition~3.1,  additional details on the experimental configuration and evaluation strategy,  and further quantitative and qualitative results that illustrate the token preservation patterns across a range of scenarios. The contents are organized as follows:
\begin{itemize}
    \item \textbf{Appendix~\ref{appendix:kde}}: Implementation details for KDE experiments in Section~3.2 of the main paper.
    \item \textbf{Appendix~\ref{appendix:Theo}}: Theoretical analysis of the proposed sensitivity estimator.
    \item \textbf{Appendix~\ref{appendix:setup}}: Experimental setup, implementation details, and evaluation protocols.
    \item \textbf{Appendix~\ref{appendix:flops}}: Further quantitative results and inference efficiency analysis. 
    \item \textbf{Appendix~\ref{appendix:visual}}: Additional qualitative results, including both successful and failure cases.
    \item \textbf{Appendix~\ref{appendix:Future_Work}}: Discussion and Future Work.
\end{itemize}

\section{Spearman Correlation Setup}
\label{appendix:kde}


For the correlation analysis in Fig.~2 of the main paper, we examined how closely the token-importance rankings from the vision encoder matched those from the projection layer. Specifically, we selected 50 random samples per dataset from MMMU and POPE. For each sample, token sensitivities were first computed using RGE at both the vision encoder output and the projection layer. To ensure stable ranking comparisons, we applied a 0.5 threshold to filter out low-sensitivity tokens before computing ranks. The Spearman’s rank correlation coefficient was then calculated for each sample, and the distribution across 50 samples was visualized using a kernel density estimate (KDE) plot.

The resulting average Spearman correlations were 0.55 for MMMU and 0.49 for POPE, indicating a consistent alignment between token rankings obtained at the projection layer and those from the full vision encoder. This confirms that the projection layer can serve as a reliable proxy for token-level importance estimation while significantly reducing the computational overhead.

\section{Theoretical Analysis}
\label{appendix:Theo}
\subsection{Proof of Proposition}

\begin{proposition}[Approximated Mean Sensitivity]\label{appendix:prop:sensitivity}
Let $M:\!\mathbb{R}^n \!\to\! \mathbb{R}^m$ be differentiable at $x\!\in\! \mathbb{R}^n$ with Jacobian $J(x)\!=\! \nabla M(x)$. 
Let $u \!\sim\!\mathcal{N}(0, I_n)$ be an isotropic Gaussian perturbation and $h\!>\! 0$ a small step size. 
Define the finite-difference sensitivity
$S(x)\!=\!\mathbb{E}_u\left[\left\|\frac{M(x+hu) - M(x-hu)}{2h}\right\|_2\right].$
Then, for sufficiently small $h$,
\begin{equation}
S(x) = \mathbb{E}_u \big[ \| J(x) u \|_2 \big] + O(h^2).
\end{equation}
\end{proposition}

\begin{proof}
Since $M$ is differentiable at $x$, we apply a first-order Taylor expansion around $x$ for perturbations $hu$:
\begin{align}
M(x + hu) &= M(x) + h J(x) u + O(h^2), \\
M(x - hu) &= M(x) - h J(x) u + O(h^2).
\end{align}
Subtracting and dividing by $2h$ gives the symmetric finite-difference approximation:
\begin{equation}
\frac{M(x+hu) - M(x-hu)}{2h} = J(x) u + O(h^2).
\end{equation}
Taking the $\ell_2$-norm,
\begin{align}
   \left\|\frac{M(x\!+\!hu)\!-\!M(x\!-\!hu)}{2h}\right\|_2 
&= \| J(x) u \!+\! O(h^2) \|_2 \\
&= \| J(x) u \|_2 \!+\! O(h^2).   
\end{align}
Finally, taking expectation over isotropic Gaussian perturbations $u \sim \mathcal{N}(0, I_n)$ yields
\begin{equation}
S(x) = \mathbb{E}_u \big[ \| J(x) u \|_2 \big] + O(h^2).
\end{equation}
\end{proof}

This proposition establishes that the finite-difference sensitivity $S(x)$, computed using small isotropic Gaussian perturbations, provides an accurate approximation of the mean local effect of input changes on the output. Specifically, for sufficiently small step size $h$, the finite-difference estimate is equivalent, up to an $O(h^2)$ error, to the expected $\ell_2$-norm of the Jacobian applied to random Gaussian directions. Intuitively, this means that $S(x)$ captures the average magnitude of output variation induced by small, randomly oriented perturbations in the input space. By sampling $u$ from an isotropic Gaussian, all directions are treated equally, ensuring an unbiased and comprehensive measure of token sensitivity without requiring backpropagation.

\section{Experimental Setup}
\label{appendix:setup}
\subsection{Model Settings} 
We evaluate the effectiveness of \ours\ on widely used VLMs, including LLaVA-v1.5-7B~\citep{LLaVA-1.5}, LLaVA-v1.5-13B~\citep{LLaVA-1.5}, and LLaVA-1.6-7B~\citep{LLaVA-NeXT} (also referred to as LLaVA-NeXT-7B), and Qwen2.5-VL-7B~\citep{Qwen2.5}. All LLaVA models adopt the CLIP~\citep{CLIP} as the vision encoder and Vicuna~\citep{vicuna} as the base language model.

LLaVA-v1.5 models process images at $336 \times 336$ resolution, yielding 576 visual tokens, while LLaVA-NeXT-7B supports higher resolutions (up to $672\times672$), generating up to 2,880 tokens and achieving a 6.0\% gain at the cost of 3.5$\times$ more computation. Qwen2.5-VL-7B, in contrast, utilizes a dynamic-resolution ViT encoder with window attention and is built upon the Qwen2.5-7B language model, supporting a variable number of visual tokens depending on input resolution. Across all experiments, our pruning is applied in a fully training-free and calibration-free manner. 

\subsection{Implementation Details} 
All experiments are conducted on 4$\times$NVIDIA A6000 GPUs with a batch size of 1. \ours is entirely training-free and attention-free, requiring no manual specification of layers in either the LMM or the vision encoder. Token selection is performed at the lightweight projection layer, which enables seamless integration across different VLM architectures. Sensitivities are also computed at this layer using simple perturbation-based operations, ensuring negligible computational overhead during inference.

For pruning ratios, we adopt 66.7\%\,/\,77.8\%\,/\,88.9\% for LLaVA-v1.5 and 77.8\%\,/\,88.9\%\,/\,94.4\% for LLaVA-NeXT-7B. In the latter case, we follow the implementation of VisionZip~\citep{visionzip}, where the model dynamically samples up to five image patches, resulting in as many as 2,880 vision tokens. For example, with a pruning budget of 160 tokens, we retain 32 tokens per patch across five patches (32 × 5 = 160). If fewer patches are sampled (\eg four), the number of retained tokens is adjusted proportionally (\eg 128 tokens for 160\,/\,2880). We applied a low-rank factorization ($k=128$) to the MM-projector layers to further boost efficiency on LLaVA-NeXT, due to the large number of visual tokens. For the dynamic-resolution Qwen2.5-VL-7B, we evaluate at 10\% and 20\% token retention rates.

Finally, as validated in ablation, our method remains robust across different hyperparameter choices. Unless otherwise noted, we fix the perturbation hyperparameters to $m=64$ and $h=0.01$ for all experiments. Evaluation is performed using the \texttt{lmms-eval}~\citep{Lmms-eval}
 framework under official protocols and metrics.

\begin{table}[t!]
\centering
\small
\vspace{-3mm}
\caption{Summary of primary evaluation metrics.}
\vspace{-3mm}
\label{tab:benchmark_metrics}
    \resizebox{\linewidth}{!}{
    \setlength{\tabcolsep}{0pt}
    \begin{tabular}{l c}
    \toprule
    \textbf{Benchmark} & \textbf{Primary Metric} \\
    \midrule
    VQAv2, GQA, ScienceQA, TextVQA & Accuracy (Acc.) \\
    MMBench, MMMU, SeedBench & Accuracy (Acc.) \\
    POPE & F1-score (F1) \\
    MME & Perception + Cognition (P+C) \\
    \bottomrule
    \end{tabular}
    }
    \vspace{-3mm}
\end{table}

\begin{table*}[t!]
\scriptsize
\vspace{-5mm}
\centering
\caption{Performance Comparison on LLaVA-1.5-7B.}
\vspace{-3mm}
\label{tab:appendix_LLaVA-1.5-7B}
\resizebox{\textwidth}{!}{
\setlength{\tabcolsep}{7pt}
\begin{tabular}{l|ccccccccc|c}
\toprule
\textbf{Method} & 
\begin{tabular}[c]{@{}c@{}} \textbf{GQA} \\ Acc. $\uparrow$ \end{tabular} & 
\begin{tabular}[c]{@{}c@{}} \textbf{MMB} \\ Acc. $\uparrow$ \end{tabular} & 
\begin{tabular}[c]{@{}c@{}} \textbf{MME} \\ P+C $\uparrow$ \end{tabular} & 
\begin{tabular}[c]{@{}c@{}} \textbf{POPE} \\ F1 $\uparrow$ \end{tabular} & 
\begin{tabular}[c]{@{}c@{}} \textbf{SQA} \\ Acc. $\uparrow$ \end{tabular} & 
\begin{tabular}[c]{@{}c@{}} \textbf{VQA$^{V2}$} \\ Acc. $\uparrow$ \end{tabular} & 
\begin{tabular}[c]{@{}c@{}} \textbf{VQA$^{Text}$} \\ Acc. $\uparrow$ \end{tabular} & 
\begin{tabular}[c]{@{}c@{}} \textbf{MMMU} \\ Acc. $\uparrow$ \end{tabular} & 
\begin{tabular}[c]{@{}c@{}} \textbf{SEED} \\ Acc. $\uparrow$ \end{tabular} & 
\begin{tabular}[c]{@{}c@{}} \textbf{Avg.} \\ $\uparrow$ \end{tabular} \\
\midrule
\rowcolor{gray!20}
\multicolumn{11}{c}{\textit{Total 576 Tokens}} \\
\midrule
LLaVA-1.5-7B & 61.90 & 64.70 & 1862.00 & 85.90 & 69.50 & 78.50 & 58.20 & 36.30 & 58.60 & 100\% \\
\midrule
\rowcolor{gray!20}
\multicolumn{11}{c}{\textit{Retain 192 Tokens} \textcolor{ForestGreen}{$\downarrow$ 66.7\%}} \\
\midrule
FastV {\scriptsize \textcolor{gray}{(ECCV 2024)}} & 52.70 & 61.20 & 1612.00 & 64.80 & 67.30 & 67.10 & 52.50 & 34.30 & 57.10 & 89.6\% \\
SparseVLM {\scriptsize \textcolor{gray}{(ICML 2025)}} & 57.60 & 62.50 & 1721.00 & 83.60 & 69.10 & 75.60 & 56.10 & 33.80 & 55.80 & 95.5\% \\
VisionZip {\scriptsize \textcolor{gray}{(CVPR 2025)}} & 59.30 & \textbf{63.00} & \textbf{1782.60} & 85.30 & 68.90 & 76.80 & 57.30 & 36.60 & 56.40 & 97.9\% \\
DivPrune {\scriptsize \textcolor{gray}{(CVPR 2025)}} & 59.97 & 62.54 & 1762.23 & 87.00 & 68.91 & 76.87 & 56.97 & 35.44 & 58.71 & 98.0\% \\
\textbf{\ours (Ours)} & \textbf{60.03} & 62.89 & 1781.66 & \textbf{87.24} & \textbf{69.16} & \textbf{77.34} & \textbf{57.30} & \textbf{36.11} & \textbf{58.80} & \textcolor[HTML]{EB4949}{\textbf{98.6}\%} \\

\midrule
\rowcolor{gray!20}
\multicolumn{11}{c}{\textit{Retain 128 Tokens} \textcolor{ForestGreen}{$\downarrow$ 77.8\%}} \\
\midrule
FastV {\scriptsize \textcolor{gray}{(ECCV 2024)}} & 49.60 & 56.10 & 1490.00 & 59.60 & 60.20 & 61.80 & 50.60 & 34.90 & 55.90 & 84.5\% \\
SparseVLM {\scriptsize \textcolor{gray}{(ICML 2025)}} & 56.00 & 60.00 & 1696.00 & 80.50 & 67.10 & 73.80 & 54.90 & 33.80 & 53.40 & 93.0\% \\
VisionZip {\scriptsize \textcolor{gray}{(CVPR 2025)}} & 57.60 & 62.00 & \textbf{1761.70} & 83.20 & 68.90 & 75.60 & 56.80 & \textbf{37.90} & 54.90 & 96.8\% \\
DivPrune {\scriptsize \textcolor{gray}{(CVPR 2025)}} & 59.25 & \textbf{62.03} & 1718.22 & 86.72 & \textbf{68.96} & 75.96 & 56.06 & 35.56 & 56.98 & 96.9\% \\
\textbf{\ours (Ours)} & \textbf{59.49} & 61.86 & 1751.60 & \textbf{87.13}	& 68.91	& \textbf{76.57} & \textbf{57.87} & 35.67 & \textbf{57.53} & \textcolor[HTML]{EB4949}{\textbf{97.8}\%} \\

\midrule
\rowcolor{gray!20}
\multicolumn{11}{c}{\textit{Retain 64 Tokens} \textcolor{ForestGreen}{$\downarrow$ 88.9\%}} \\
\midrule
FastV {\scriptsize \textcolor{gray}{(ECCV 2024)}}& 46.10 & 48.00 & 1256.00 & 48.00 & 51.10 & 55.00 & 47.80 & 34.00 & 51.90 & 75.5\% \\
SparseVLM {\scriptsize \textcolor{gray}{(ICML 2025)}}  & 52.70 & 56.20 & 1505.00 & 75.10 & 62.20 & 68.20 & 51.80 & 32.70 & 51.10 & 87.0\% \\
VisionZip {\scriptsize \textcolor{gray}{(CVPR 2025)}} & 55.10 & 60.10 & \textbf{1690.00} & 77.00 & \textbf{69.00} & 72.40 & \textbf{55.50} & 36.20 & 52.20 & 93.1\% \\
DivPrune {\scriptsize \textcolor{gray}{(CVPR 2025)}} & 57.78 & 59.28 & 1674.40 & 85.56 & 68.17 & 74.11 & 54.69 & \textbf{35.56} & 55.13 & 94.8\% \\
\textbf{\ours (Ours)} & \textbf{58.47} & \textbf{60.22} & 1675.59 & \textbf{85.86} & 68.27 & \textbf{75.02} & 55.35 & 35.44 & \textbf{55.84} & \textcolor[HTML]{EB4949}{\textbf{95.5}\%} \\

\bottomrule
\bottomrule
\end{tabular}
\vspace{-5mm}
}
\end{table*}

\subsection{Evaluation Protocol and Benchmark Datasets}
We conduct a comprehensive evaluation of \ours\ across \textbf{nine widely adopted vision-language benchmarks}, spanning four core capabilities: \textit{Visual Question Answering}, \textit{Advanced Multimodal Reasoning}, \textit{Object Hallucination Evaluation}, and \textit{Comprehensive Multimodal Assessment}. All experiments strictly follow the official evaluation protocols, metrics, and data splits of each benchmark to ensure fair and reproducible comparisons.

To facilitate a unified and interpretable comparison, we report both per-benchmark scores and a normalized average performance (Avg.), computed as the mean relative score across benchmarks with respect to the unpruned baseline.
Depending on the benchmark, we report Accuracy (Acc), F1-score (F1), or Perception+Cognition (P+C), summarized in Table~\ref{tab:benchmark_metrics}. All evaluations of \ours are performed under a single-model, zero-shot setting, without any task-specific fine-tuning.

\vspace{-2mm}
\paragraph{Visual Question Answering (VQA).}
This category evaluates a model’s ability to ground language understanding in visual content. Performance across all VQA benchmarks is measured by \textbf{Accuracy (Acc)}. We select four representative benchmarks covering diverse scenarios:
\begin{itemize}
    \item \textbf{VQAv2-Test-Dev}~\citep{VQAV2}: General-purpose VQA with real-world images and open-ended questions.
    \item \textbf{GQA}~\citep{GQA}: Focused on compositional reasoning over scene graphs and structured images.
    \item \textbf{ScienceQA (IMG)}~\citep{scienceqa}: Multimodal science questions requiring domain knowledge and diagram interpretation.
    \item \textbf{TextVQA}~\citep{textvqa}: Requires OCR capabilities to reason over text embedded within images.
\end{itemize}

\vspace{-2mm}
\paragraph{Advanced Multimodal Reasoning.}
To probe deeper reasoning capacities beyond standard VQA, we evaluate on three challenging benchmarks. Performance on these benchmarks is also measured by \textbf{Accuracy (Acc)}:
\begin{itemize}
    \item \textbf{MMBench}~\citep{mmbench}: Assesses perception and reasoning across 20 fine-grained skill areas.
    \item \textbf{MMMU}~\citep{mmmu}: Requires expert-level multimodal reasoning across 30+ subjects grouped into six major disciplines (\eg Art \& Design, Science, Engineering, Medicine), often involving complex diagrams and charts
    \item \textbf{SeedBench}~\citep{Seedbench}: Designed for evaluating multimodal large language models across diverse visually grounded question types, with an emphasis on perception, reasoning, and knowledge.
\end{itemize}

\vspace{-2mm}
\paragraph{Object Hallucination Evaluation.}
To quantify the critical failure mode of object hallucination, we adopt the \textbf{POPE}~\citep{POPE} benchmark, which measures factuality in object recognition through binary existence questions. Performance is evaluated using the \textbf{F1-score (F1)} over object existence predictions, balancing precision and recall to reflect grounding reliability.

\vspace{-2mm}
\paragraph{Comprehensive Multimodal Assessment.}
For a holistic evaluation of both perceptual and cognitive abilities across numerous sub-tasks (\eg OCR, counting, attribute recognition), we employ the \textbf{MME}~\citep{mme}. It reports separate scores for Perception and Cognition tasks, summed to form the combined \textbf{Perception and Cognition score (P+C)}.



\begin{table*}[t!]
\scriptsize
\vspace{-5mm}
\centering
\caption{Performance Comparison on LLaVA-1.5-13B.}
\vspace{-3mm}
\label{tab:appendix_LLaVA-1.5-13B}
\resizebox{\textwidth}{!}{
\setlength{\tabcolsep}{7 pt}
\begin{tabular}{l|ccccccccc|c} 
\toprule
\textbf{Method} & 
\begin{tabular}[c]{@{}c@{}} \textbf{GQA} \\ Acc. $\uparrow$ \end{tabular} & 
\begin{tabular}[c]{@{}c@{}} \textbf{MMB} \\ Acc. $\uparrow$ \end{tabular} & 
\begin{tabular}[c]{@{}c@{}} \textbf{MME} \\ P+C $\uparrow$ \end{tabular} & 
\begin{tabular}[c]{@{}c@{}} \textbf{POPE} \\ F1 $\uparrow$ \end{tabular} & 
\begin{tabular}[c]{@{}c@{}} \textbf{SQA} \\ Acc. $\uparrow$ \end{tabular} & 
\begin{tabular}[c]{@{}c@{}} \textbf{VQA$^{V2}$} \\ Acc. $\uparrow$ \end{tabular} & 
\begin{tabular}[c]{@{}c@{}} \textbf{VQA$^{Text}$} \\ Acc. $\uparrow$ \end{tabular} & 
\begin{tabular}[c]{@{}c@{}} \textbf{MMMU} \\ Acc. $\uparrow$ \end{tabular} & 
\begin{tabular}[c]{@{}c@{}} \textbf{SEED-I} \\ Acc. $\uparrow$ \end{tabular} & 
\begin{tabular}[c]{@{}c@{}} \textbf{Avg.} \\ $\uparrow$ \end{tabular} \\

\midrule
\rowcolor{gray!20}
\multicolumn{11}{c}{\textit{Total 576 Tokens}} \\
\midrule
LLaVA-1.5-13B & 63.20 & 67.70 & 1818.00 & 85.90 & 72.80 & 80.00 & 61.30 & 36.40 & 66.90 & 100\% \\
\midrule
\rowcolor{gray!20}
\multicolumn{11}{c}{\textit{Retain 192 Tokens} \textcolor{ForestGreen}{$\downarrow$ 66.7\%}} \\
\midrule
VisionZip {\scriptsize \textcolor{gray}{(CVPR 2025)}} & 59.10 & \textbf{66.90} & 1754.00 & 85.10 & \textbf{73.50} & 78.10 & \textbf{59.50} & 36.40 & 65.20 & 97.9\% \\
DivPrune {\scriptsize \textcolor{gray}{(CVPR 2025)}} & 59.42 & 66.58 & \textbf{1781.50} & \textbf{86.76} & 72.88 & 77.98 & 58.46 & 36.56 & \textbf{65.72} & 98.1\% \\

\textbf{\ours (Ours)} & \textbf{59.95} & 66.67 & 1762.41 & 86.73 & 73.12 & \textbf{78.65} & 59.11	& \textbf{37.33}	& 65.56 & \textcolor[HTML]{EB4949}{\textbf{98.6}}\% \\

\midrule
\rowcolor{gray!20}
\multicolumn{11}{c}{\textit{Retain 128 Tokens} \textcolor{ForestGreen}{$\downarrow$ 77.8\%}} \\
\midrule
VisionZip {\scriptsize \textcolor{gray}{(CVPR 2025)}} & 57.90 & 66.70 & 1743.00 & 85.20 & \textbf{74.00} & 76.80 & 58.70 & \textbf{36.10} & 63.80 & 97.0\% \\
DivPrune {\scriptsize \textcolor{gray}{(CVPR 2025)}} & 58.89 & 66.07 & 1748.56 & 86.53 & 72.83 & 77.10 & 58.17 & 35.56 & 64.22 & 97.0\% \\

\textbf{\ours (Ours)} & \textbf{58.89} & \textbf{67.01} & \textbf{1791.10}	& \textbf{86.95} & 73.38 & \textbf{77.83} & \textbf{58.80} & 35.56 & \textbf{64.50} & \textcolor[HTML]{EB4949}{\textbf{97.8}}\% \\

\midrule
\rowcolor{gray!20}
\multicolumn{11}{c}{\textit{Retain 64 Tokens} \textcolor{ForestGreen}{$\downarrow$ 88.9\%}} \\
\midrule
VisionZip {\scriptsize \textcolor{gray}{(CVPR 2025)}} & 56.20 & \textbf{64.90} & 1676.00 & 76.00 & \textbf{74.40} & 73.70 & 57.40 & \textbf{36.40} & 60.40 & 93.7\% \\
DivPrune {\scriptsize \textcolor{gray}{(CVPR 2025)}} & 57.66 & 64.60 & \textbf{1777.93} & 84.80 & 71.34 & 75.20 & 57.11 & 35.22 & 62.44 & 95.4\% \\

\textbf{\ours (Ours)} & \textbf{58.58} & 64.78 & \textbf{1780.03} & \textbf{85.34} & 72.09 & \textbf{76.39} & \textbf{58.59 }& 36.00 & \textbf{63.02} & \textcolor[HTML]{EB4949}{\textbf{96.5}}\% \\

\bottomrule
\bottomrule
\end{tabular}
}
\vspace{-3mm}
\end{table*}
\begin{figure*}[t]
    \centering
    \setlength{\abovecaptionskip}{1mm}
    \begin{subfigure}[b]{0.49\linewidth} 
        \includegraphics[width=\textwidth]{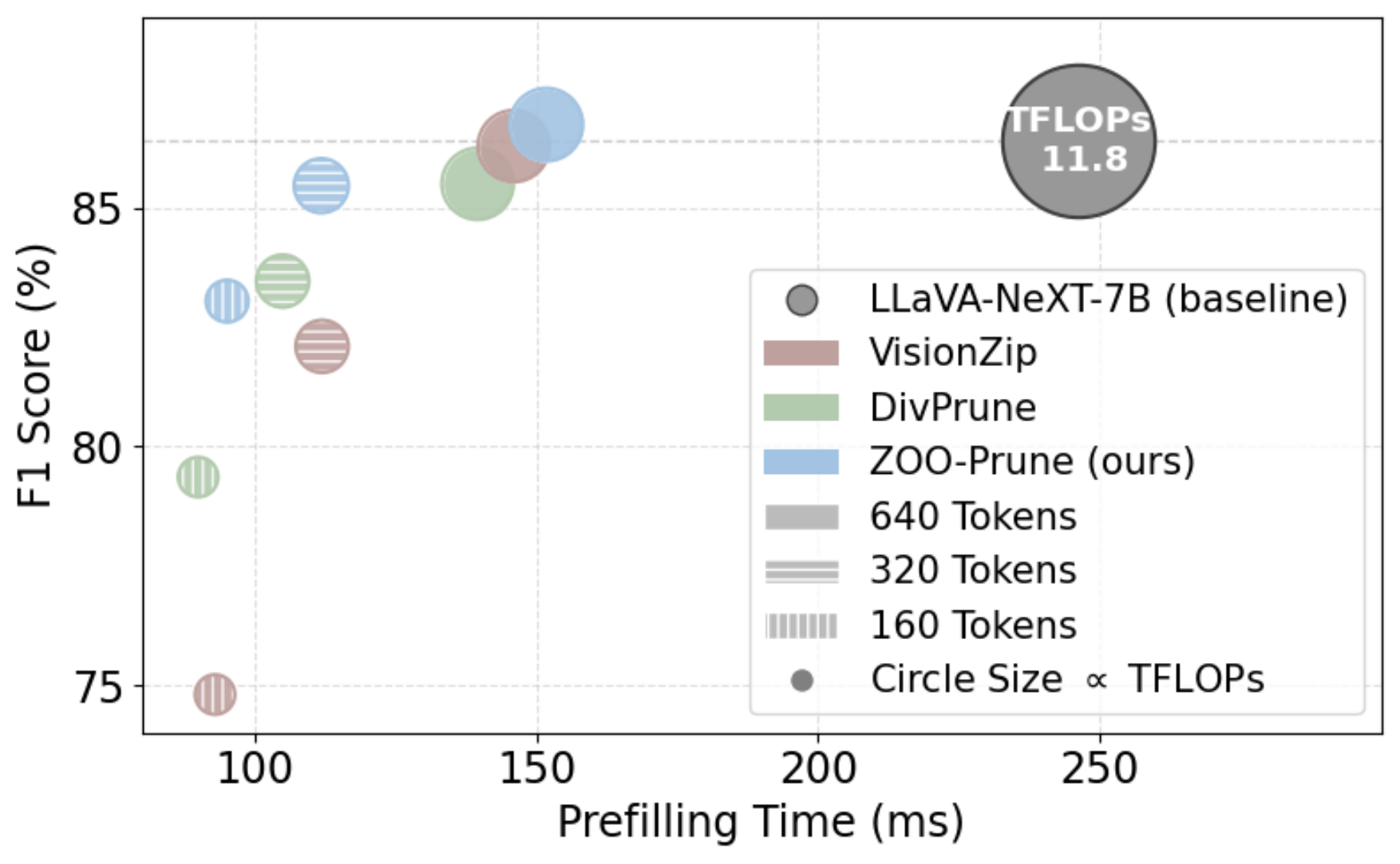}
        \caption{Prefilling Time (ms) vs. Performance (\%) vs. TFLOPs}
        \vspace{1mm}
        \label{fig:prefilling}
    \end{subfigure}
    \begin{subfigure}[b]{0.49\linewidth}
        \includegraphics[width=\textwidth]{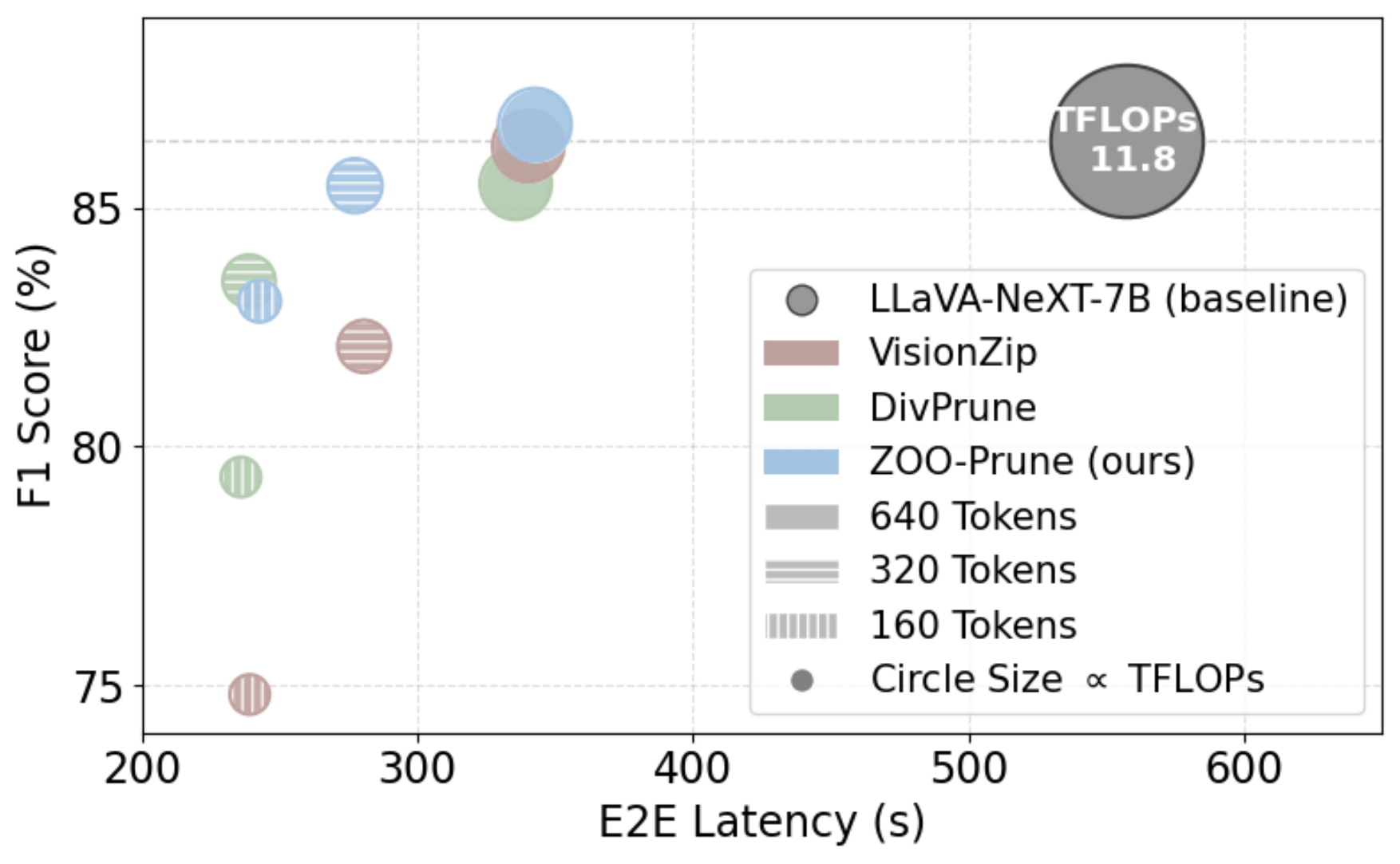}
        \caption{E2E Latency (secs) vs. Performance (\%) vs. TFLOPs}
        \vspace{1mm}
        \label{fig:latency}
    \end{subfigure}
    \caption{
    Inference efficiency on the POPE benchmark relative to the LLaVA-NeXT-7B baseline.
    Circle size is proportional to TFLOPs. \ours achieves the strongest accuracy–efficiency trade-off, sustaining higher POPE performance while operating at substantially lower latency and computational cost.
    }
    \label{fig:appdix_Inference_efficiency}
    \vspace{-2mm}
\end{figure*}

\section{Further Quantitative Results and Analysis}
\label{appendix:flops}
\subsection{Additional Quantitative Results}

\paragraph{Results on LLaVA-1.5-7B:}
Table~\ref{tab:appendix_LLaVA-1.5-7B} reports the results on LLaVA-1.5-7B, the most widely used model in the LLaVA family. Across all token pruning levels, \ours achieves the strongest overall performance. With a 66.7\% pruning (192 tokens), it reaches an $\text{Avg.}$ of 98.6\%, surpassing the previous best method VisionZip (97.9\%) and achieving leading scores on challenging tasks such as MMMU (36.11\%) and SEED (58.80\%).
As pruning becomes more aggressive, the advantage of \ours becomes even clearer. At 128 tokens, it maintains the best average score of 97.8\%, substantially outperforming FastV (84.5\%) and SparseVLM (93.0\%). \ours also delivers top results on SQA (59.49\%), POPE (87.13\%), VQAv2 (76.57\%), and TextVQA (57.87\%).
Even under the extremely compressed 64-token setting, \ours retains an average of 95.5\%, outperforming the attention-based VisionZip by +2.4\% and the diversity-based DivPrune by +0.7\%. These results demonstrate that our sensitivity-aware diversity selection reliably preserves key visual cues and remains robust even under aggressive pruning.

\vspace{-2mm}
\paragraph{Results on LLaVA-1.5-13B:}
The robustness of ZOO-Prune is further validated in Table~\ref{tab:appendix_LLaVA-1.5-13B}, where methods are applied to the larger LLaVA-1.5-13B model. This evaluation tests the ability of pruning strategies to generalize to higher-capacity backbones. With 192 tokens retained, \ours again obtains the highest average score (98.6\%).
Under the challenging 64-token setting, \ours shows a clear performance advantage. It reaches 96.5\% average accuracy, outperforming VisionZip by +2.8\% and DivPrune by +1.1\%. Retaining strong performance while discarding 88.9\% of visual tokens highlights the robustness of our approach.
Overall, across both 7B and 13B variants, ZOO-Prune consistently provides the best accuracy–efficiency trade-off, demonstrating strong generalization to different model capacities and pruning regimes.

\vspace{-2mm}
   
\begin{table*}[t!]
\scriptsize
\vspace{6mm}
\centering
\caption{
Performance comparison between the projector and vision encoder layers as sensitivity proxies.
The projector consistently provides the strongest accuracy across pruning ratios.
}
\vspace{-1mm}
\label{tab:LLaVA-NeXT-encoder-layer}
\resizebox{\textwidth}{!}{
\setlength{\tabcolsep}{7 pt}
\begin{tabular}{l|cccccccc|c} 
\toprule
\textbf{Vision Encoder} & 
\begin{tabular}[c]{@{}c@{}} \textbf{GQA} \\ Acc. $\uparrow$ \end{tabular} & 
\begin{tabular}[c]{@{}c@{}} \textbf{MMB} \\ Acc. $\uparrow$ \end{tabular} & 
\begin{tabular}[c]{@{}c@{}} \textbf{MME} \\ P+C $\uparrow$ \end{tabular} & 
\begin{tabular}[c]{@{}c@{}} \textbf{POPE} \\ F1 $\uparrow$ \end{tabular} & 
\begin{tabular}[c]{@{}c@{}} \textbf{SQA} \\ Acc. $\uparrow$ \end{tabular} & 
\begin{tabular}[c]{@{}c@{}} \textbf{VQA$^{Text}$} \\ Acc. $\uparrow$ \end{tabular} & 
\begin{tabular}[c]{@{}c@{}} \textbf{MMMU} \\ Acc. $\uparrow$ \end{tabular} & 
\begin{tabular}[c]{@{}c@{}} \textbf{SEED-I} \\ Acc. $\uparrow$ \end{tabular} & 
\begin{tabular}[c]{@{}c@{}} \textbf{Avg.} \\ $\uparrow$ \end{tabular} \\
\midrule
\rowcolor{gray!20}
\multicolumn{10}{c}{\textit{Total 2880 Tokens}} \\
\midrule
LLaVA-NeXT-7B & 64.20 & 67.90 & 1842.00 & 86.40 & 70.20 & 61.30 & 35.10 & 70.20 & 100\% \\
\midrule
\rowcolor{gray!20}
\multicolumn{10}{c}{\textit{Retain 640 Tokens} \textcolor{mygreen}{$\downarrow$ 77.8\%}} \\
\midrule
        Layer 01 & 61.23 & 65.29 & 1779.27 & 81.41 & 67.87 & 52.49 & 37.56 & 66.25 & 95.8\% \\
        Layer 03 & 61.96 & \textbf{65.64} & 1819.21 & 86.35 & \textbf{68.47} & 57.80 & 36.67 & 67.68 & 98.1\% \\
        Layer 05 & 61.73 & 65.55 & 1815.92 & 86.74 & 68.17 & 57.43 & 36.78 & 67.59 & 97.9\% \\
        Layer 07 & 61.85 & 65.29 & 1815.84 & 86.41 & 67.18 & 56.85 & 37.89 & 68.07 & 98.1\% \\
        Layer 09 & 62.18 & 65.12 & 1813.60 & 86.44 & 67.97 & 57.39 & 37.67 & 67.33 & 98.1\% \\
        Layer 11 & 61.98 & 65.21 & 1806.98 & 86.41 & 68.17 & 57.41 & 37.22 & 67.69 & 98.0\% \\
        Layer 13 & 61.93 & 65.12 & 1803.63 & 86.47 & 67.72 & 54.06 & 37.78 & 67.62 & 97.4\% \\
        Layer 15 & 62.03 & 65.38 & 1782.60 & 86.72 & 67.92 & 56.41 & 37.56 & 67.98 & 97.8\% \\
        Layer 17 & 61.99 & 65.38 & 1803.06 & \textbf{86.83} & 67.87 & 55.57 & \textbf{38.00} & \textbf{68.28} & 98.0\% \\
        Layer 19 & 61.94 & 65.03 & \textbf{1840.15} & 86.78 & 67.97 & 56.07 & 37.44 & 68.12 & 98.1\% \\
        Layer 21 & 62.03 & 65.29 & 1820.98 & 86.78 & 67.77 & 57.94 & 36.89& 67.71 & 98.1\% \\
        Layer 23 & 61.70 & 65.46 & 1829.87 & 84.83 & 67.23 & 56.53 & 37.56 & 67.32 & 97.6\% \\
        Layer 24 & 61.99 & 65.21 & 1838.67 & 86.34 & 67.63 & \textbf{58.31} & 36.56 & 67.77 & 98.1\% \\
\textbf{Projector (Ours)} & \textbf{62.19} & 65.21 & 1816.45 & 86.75 & 68.02 & 57.98 & 36.89 & 67.95 & \textcolor[HTML]{EB4949}{\textbf{98.2}}\% \\

\midrule
\rowcolor{gray!20}
\multicolumn{10}{c}{\textit{Retain 320 Tokens} \textcolor{mygreen}{$\downarrow$ 88.9\%}} \\
\midrule
        Layer 01 & 59.11 & 63.92 & 1664.19 & 76.08 & \textbf{68.86} & 49.38 & 36.89 & 63.94 & 92.4\% \\
        Layer 03 & 60.71 & 64.09 & 1796.69 & 84.08 & 67.87 & 56.74 & 36.44 & 65.64 & 96.3\% \\
        Layer 05 & 60.70 & 64.60 & 1789.64 & 84.48 & 67.82 & 55.77 & \textbf{38.44} & 65.77 & 96.9\% \\
        Layer 07 & 61.05 & 64.60 & 1746.22 & 83.23 & 67.67 & 55.36 & 37.11 & 65.87 & 96.0\% \\
        Layer 09 & 60.90 & 63.92 & 1798.27 & 83.74 & 67.53 & 55.81 & 37.00 & 65.62 & 96.2\% \\
        Layer 11 & 60.73 & 64.26 & 1784.26 & 84.36 & 65.80 & 56.13 & 36.56 & 65.80 & 95.9\% \\
        Layer 13 & 60.78 & 63.83 & 1712.90 & 83.76 & 67.38 & 52.13 & 36.67 & 65.64 & 94.7\% \\
        Layer 15 & 60.99 & 63.57 & 1754.49 & 84.44 & 67.53 & 55.02 & 37.22 & 66.39 & 96.0\% \\
        Layer 17 & 60.83 & \textbf{64.86} & 1799.67 & 85.00 & 67.33 & 53.70 & 37.11 & \textbf{66.48} & 96.3\% \\
        Layer 19 & 61.05 & 64.52 & \textbf{1801.79} & 85.44 & 67.63 & 54.51 & 37.67 & 66.42 & 96.8\% \\
        Layer 21 & 61.00 & 64.26 & 1764.22 & 84.37 & 66.93 & 56.05 & 37.67 & 65.72 & 96.4\% \\
        Layer 23 & 60.43 & 63.57 & 1766.28 & 81.46 & 68.12 & 54.95 & 36.89 & 65.19 & 95.3\% \\
        Layer 24 & \textbf{61.15} & 64.78 & 1775.97 & 84.23 & 67.03 & 56.14 & 37.00 & 66.29 & 96.4\% \\
\textbf{Projector (Ours)} & 60.97 & \textbf{64.86} & 1787.68 & \textbf{85.47} & 67.77 & \textbf{57.28} & 37.00 & 66.47 & \textcolor[HTML]{EB4949}{\textbf{97.1}}\%\\

\midrule
\rowcolor{gray!20}
\multicolumn{10}{c}{\textit{Retain 160 Tokens} \textcolor{mygreen}{$\downarrow$ 94.4\%}} \\
\midrule
        Layer 01 & 57.62 & 61.34 & 1590.15 & 70.07 & 68.12 & 46.57 & 36.67 & 61.45 & 89.1\% \\
        Layer 03 & 58.88 & 63.14 & 1689.52 & 79.73 & 67.13 & 55.26 & 35.67 & 62.99 & 93.2\% \\
        Layer 05 & 59.69 & 62.54 & 1705.96 & 80.85 & 68.07 & 54.47 & 36.11 & 63.99 & 93.9\% \\
        Layer 07 & 59.43 & 63.92 & 1715.39 & 78.50 & 67.38 & 52.76 & \textbf{39.00} & 63.66 & 94.3\% \\
        Layer 09 & 59.58 & 62.46 & 1712.23 & 79.35 & 67.63 & 53.39 & 35.67 & 63.42 & 93.1\% \\
        Layer 11 & 59.18 & 63.23 & 1700.83 & 80.18 & 63.59 & 54.22 & 37.22 & 63.59 & 93.3\% \\
        Layer 13 & 59.33 & 62.80 & 1690.78 & 79.55 & 67.77 & 49.61 & 38.00 & 63.60 & 93.1\% \\
        Layer 15 & 59.78 & 63.14 & 1685.21 & 82.00 & 66.98 & 52.99 & 36.44 & 64.51 & 93.8\% \\
        Layer 17 & \textbf{60.06} & 63.32 & 1712.95 & 82.54 & 66.68 & 50.78 & 37.44 & \textbf{64.78} & 94.0\% \\
        Layer 19 & 59.90 & 62.63 & 1724.21 & 82.65 & 67.63 & 51.40 & 37.56 & 64.43 & 94.2\% \\
        Layer 21 & 59.53 & 63.66 & 1720.90 & 80.69 & 67.72 & 53.94 & 36.78 & 63.46 & 94.1\% \\
        Layer 23 & 58.80 & 62.71 & 1665.75 & 75.96 & 68.17 & 52.13 & 36.67 & 62.43 & 92.2\% \\
        Layer 24 & 59.57 & 63.14 & 1714.08 & 81.01 & 67.67 & 55.00 & 37.00 & 63.87 & 94.4\% \\
\textbf{Projector (Ours)} & 59.93 & \textbf{64.18} & \textbf{1738.64} & \textbf{83.05} & \textbf{68.42} & \textbf{55.42} & 37.11 & 64.05 & \textcolor[HTML]{EB4949}{\textbf{95.4}}\%\\

\bottomrule
\bottomrule
\end{tabular}
\vspace{4mm}
}
\end{table*}

\subsection{Further Inference Efficiency Analysis}

We evaluate the computational benefits of \ours by measuring prefilling time, end-to-end (E2E) latency, and FLOPs on LLaVA-NeXT-7B using the POPE benchmark.
The total computational cost can be expressed as:
\begin{equation}
\mathrm{FLOPs}_{\mathrm{total}}
=
\mathrm{FLOPs}_{\mathrm{prefill}}(\hat n)
+
\Delta \mathrm{FLOPs}_{\mathrm{ZOO}},
\end{equation}
where $n$ and $\hat{n}$ denote the sequence length before and after pruning, respectively.
Following prior work~\citep{FastV,visionzip,Divprune}, the prefilling FLOPs of an $L$-layer LLM with hidden size $d$ and FFN intermediate size $m$ scale as:
$L(4\hat nd^2 + 2\hat n^2d + 2\hat ndm)$. 
Our pruning overhead comes only from ZOO sensitivity estimation at the lightweight projector. With $m_z$ random perturbation directions and a rank-$k$ factorization, the additional cost is:
\begin{align}
\Delta \mathrm{FLOPs}_{\mathrm{ZOO}}
&= \mathrm{FLOPs}_{\mathrm{ZOO,LR}} - \mathrm{FLOPs}_{\mathrm{proj}}\\
&=2 m_z \, n \, \frac{d}{k} (D_v + d) -nD_v d \\
&=  n d^2,  \quad \text{when }\space m_z{=}64, k{=}128.
\end{align}
Therefore, under our default configuration, this overhead is about $0.3\%$ of the baseline prefilling FLOPs.

As shown in Fig.~\ref{fig:appdix_Inference_efficiency}, all pruning methods provide noticeable efficiency gains across token budgets, but \ours offers the most favorable balance between speed and accuracy. At the most aggressive setting (160 tokens), prefilling becomes 2.59$\times$ faster and E2E latency is reduced by 2.30$\times$ relative to the baseline. FLOPs show a similarly substantial reduction: from 11.8 TFLOPs in the baseline to just 0.9 TFLOPs with \ours. Importantly, despite this nearly 13$\times$ computational reduction, \ours retains 96.12\% of the baseline’s performance (83.05\% F1), indicating only minimal accuracy degradation. These results support that \ours delivers a consistently superior accuracy–efficiency trade-off, making it a practical and effective pruning strategy for real-world VLM deployment.

\begin{figure*}[t]
\vspace{-3mm}
    \centering
    \setlength{\abovecaptionskip}{0.0cm}
    \includegraphics[width=0.95\linewidth]{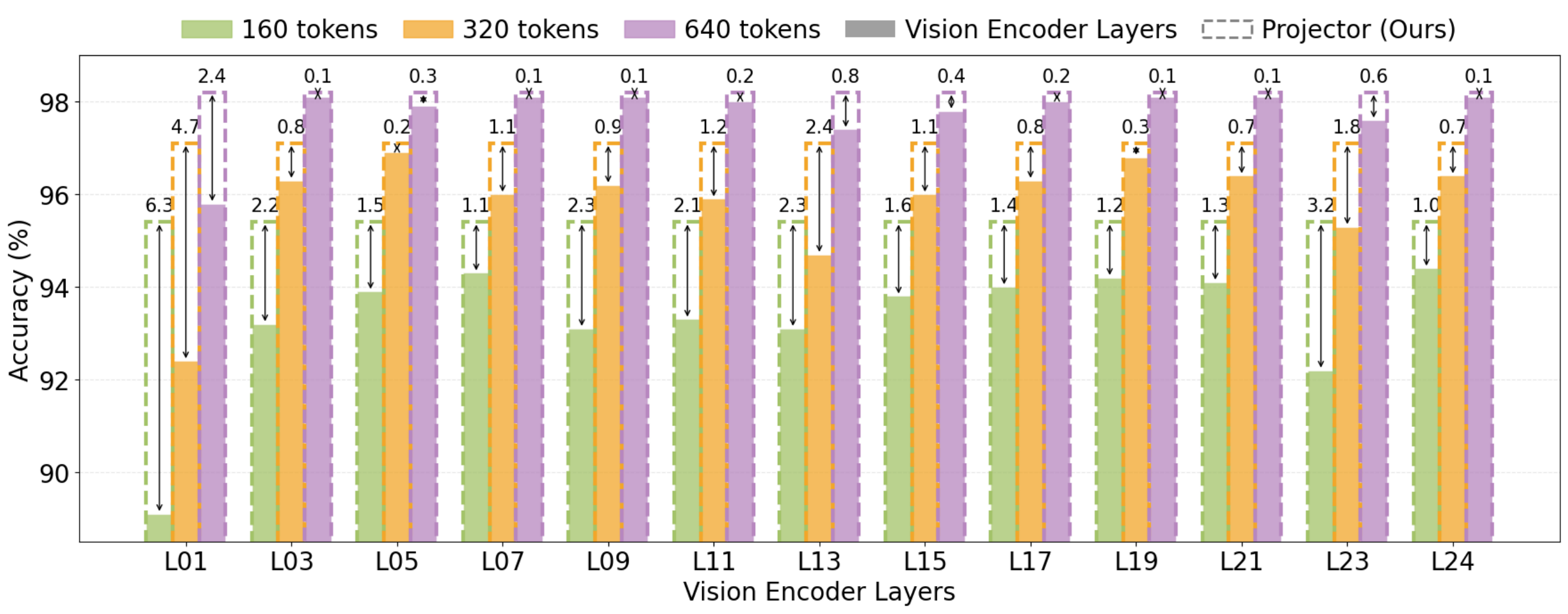}
    \caption{
    Revisiting Token Sensitivity Proxy: Projector vs. Encoder. Compared to the visual encoder, the projector is a reliable proxy for zoo-based token sensitivity estimation.
    }
    \label{fig:app: vision_encoder}
    \vspace{-3mm}
\end{figure*}

\subsection{Visual Sensitivity: Projector vs. Encoder}

The projection layer acts as a {modality-alignment bottleneck, where high-level visual features are consolidated and mapped into the LLM embedding space. Tokens that remain influential at this stage are therefore those most critical for downstream language generation. Motivated by this, \ours computes ZOO-based sensitivity at the projector rather than across the full vision encoder, avoiding costly end-to-end perturbations.

To evaluate whether the projector serves as a reliable proxy, we compare performance when sensitivity is computed at each encoder layer. As shown in Table~\ref{tab:LLaVA-NeXT-encoder-layer}, mid-to-deep encoder layers (\eg 17–24 layers) provide moderately strong estimates but consistently underperform the projector, while early layers degrade sharply under aggressive pruning (\eg Layer~1 falls to 89.1\% at a 160-token budget). Fig.~\ref{fig:app: vision_encoder} further highlights this trend: the projector achieves the highest and most stable accuracy across all pruning levels, with its advantage growing more pronounced as token budgets shrink.
Overall, these findings show that the projection layer delivers the most stable, accurate, and efficient signal for token sensitivity, motivating its use as the default estimation strategy in \ours.

\begin{figure*}[t]
    \centering
    \setlength{\abovecaptionskip}{2mm} 
    \begin{subfigure}[b]{\textwidth}
        \includegraphics[width=\linewidth]{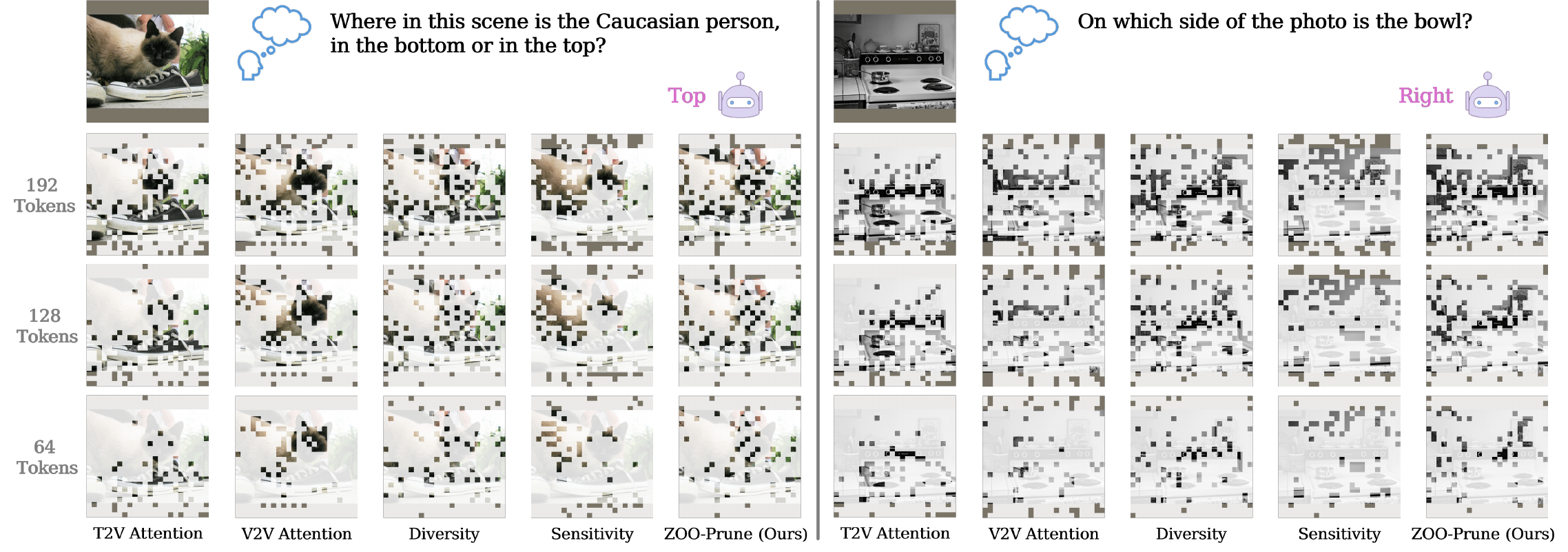}
        \label{fig:visualization_app_1}
        \vspace{2mm} 
    \end{subfigure}
    \begin{subfigure}[b]{\textwidth}
        \includegraphics[width=\linewidth]{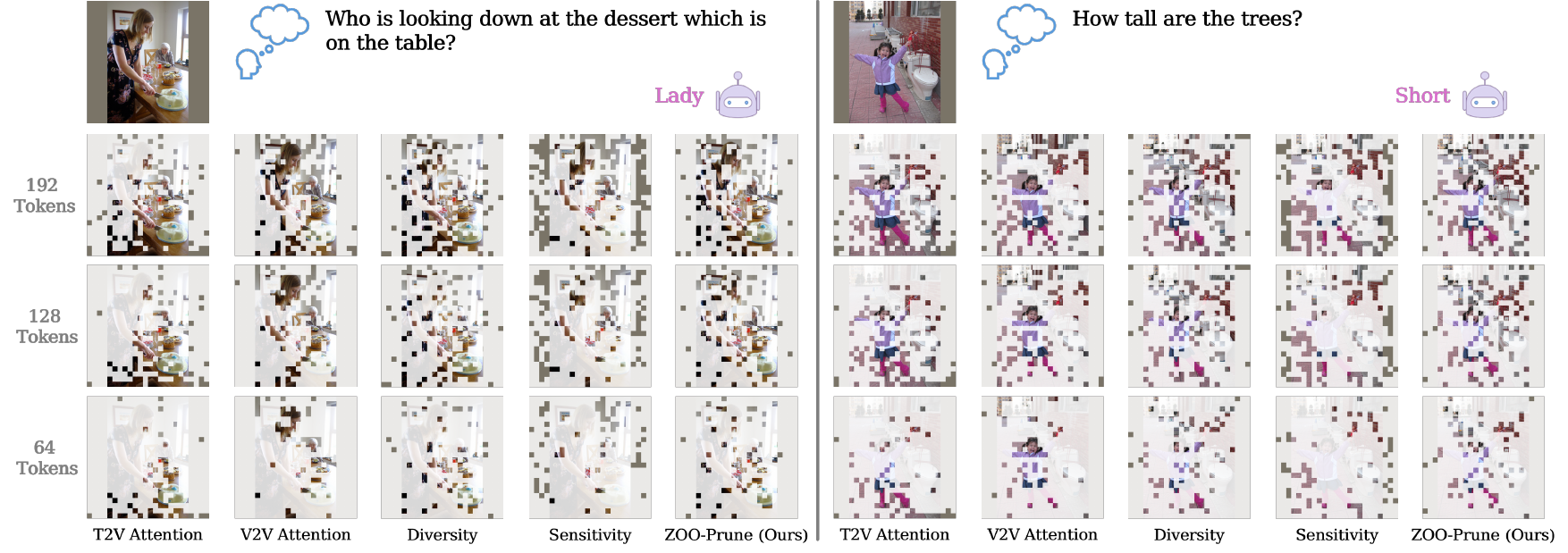}
        \label{fig:visualization_app_2}
        \vspace{2mm}
    \end{subfigure}
    \begin{subfigure}[b]{\textwidth}
    \includegraphics[width=\linewidth]{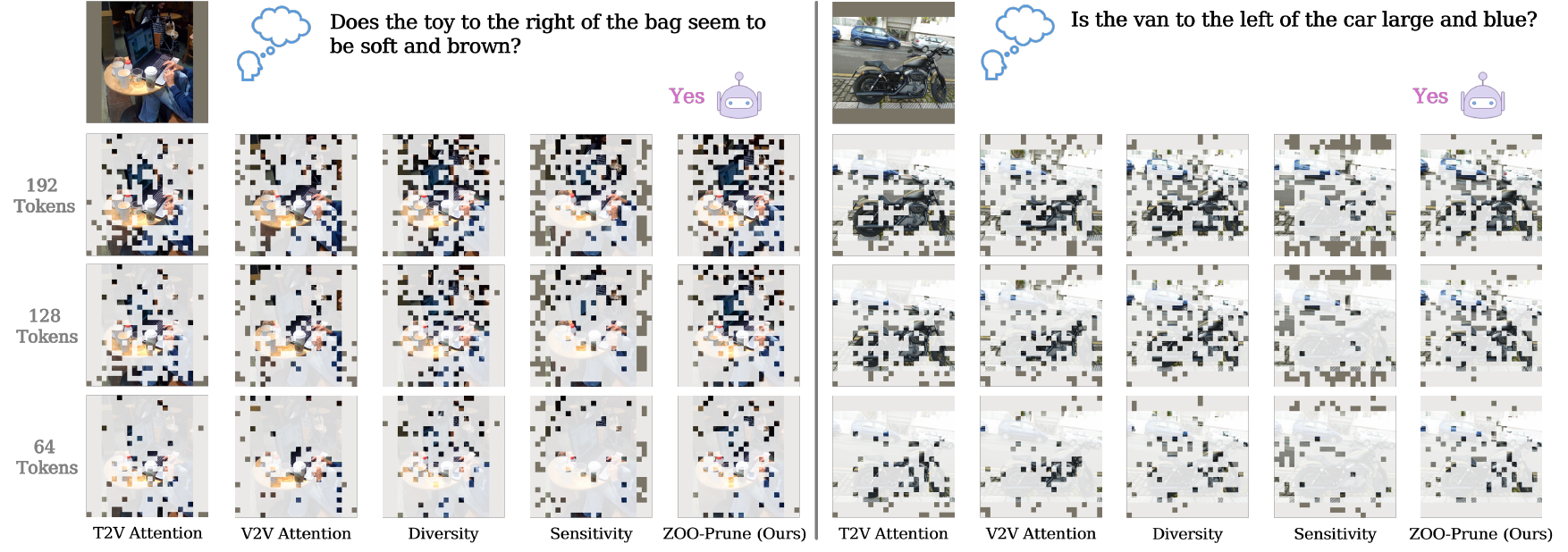}
    \label{fig:visualization_app_3}
    \end{subfigure}
    \caption{
    Selected visual tokens comparison on the GQA benchmark. Token pruning driven by text-visual (T2V) attention often suffers from positional bias, while visual-visual (V2V) attention tends to retain redundant token clusters.
    Diversity-based pruning spreads tokens broadly but lacks semantic focus. ZOO-based sensitivity can capture output-related tokens but overlooks spatial coverage. Our ZOO-Prune jointly optimizes sensitivity and diversity for balanced selection across compression ratios.
    }
    \label{fig:appdix_visualization_att}
\end{figure*}

\begin{figure*}[!t]
    \centering
    \setlength{\abovecaptionskip}{0.12cm}
    \includegraphics[width=0.994\linewidth]{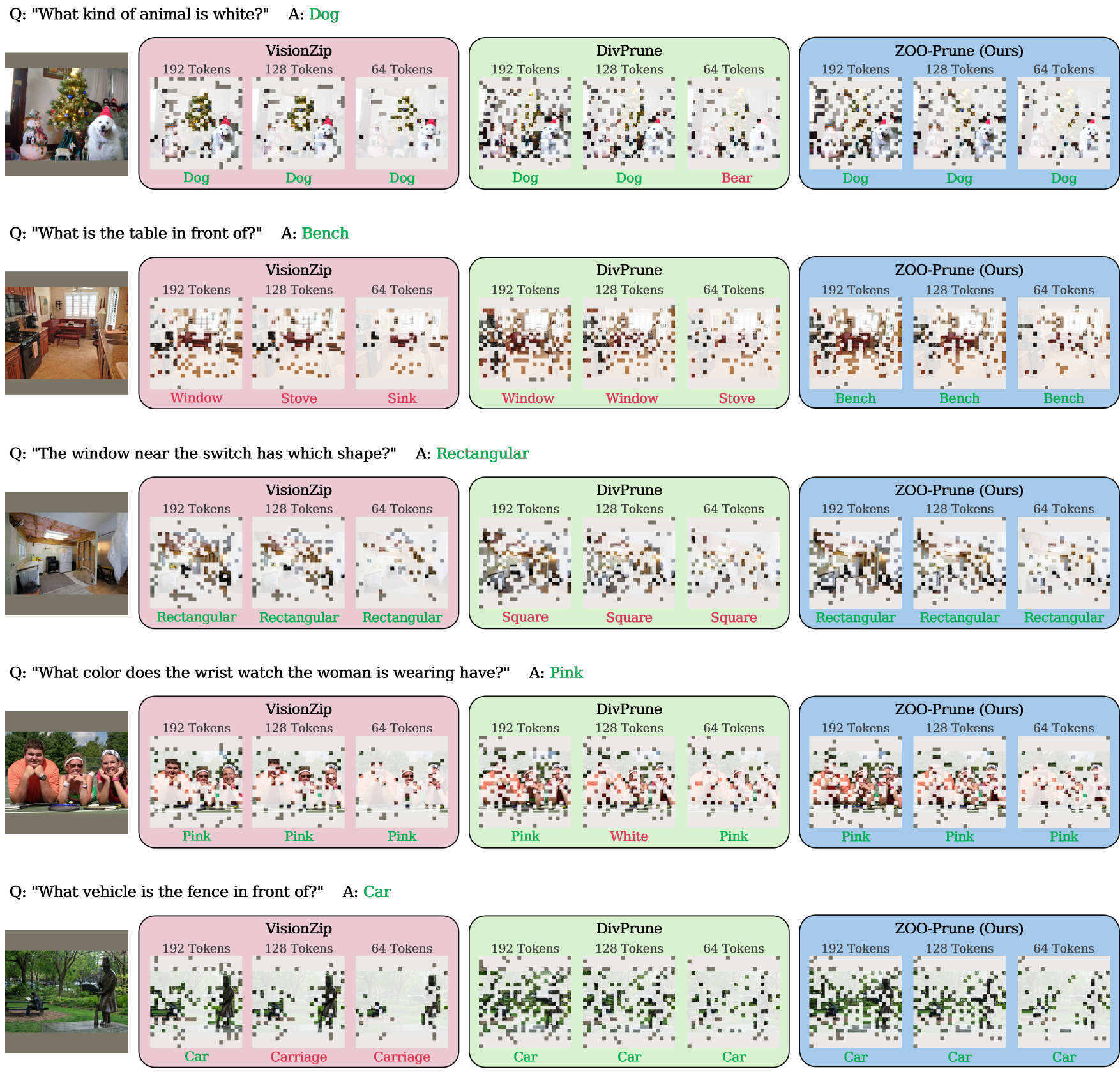}
\caption{
Qualitative comparison of pruned token masks from VisionZip, DivPrune and \ours on the GQA benchmark.
}
    \label{fig:appendix_visualization_sota}
\end{figure*}

\begin{figure*}[!t]
    \centering
    \setlength{\abovecaptionskip}{0cm}
    \includegraphics[width=\linewidth]{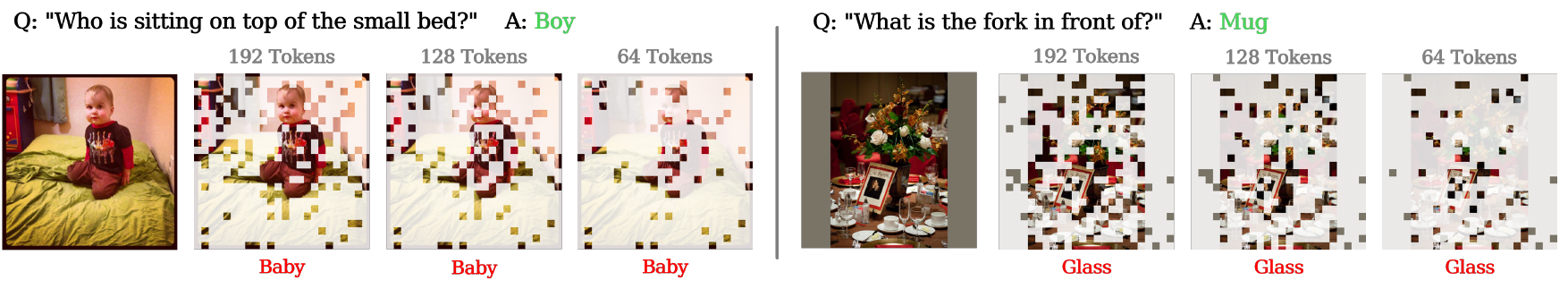}
\caption{
Failure cases of \ours illustrating two typical error patterns. Left: Predictions remain semantically close to references but differ in fine-grained age distinctions (e.g., ``boy" vs. ``baby"). Right: In visually cluttered scenes with multiple objects, the model makes incorrect predictions due to confusion among different items present in the scene (e.g., ``mug" vs. ``glass").
}
    \label{fig:app:FailedCase}
\end{figure*}

\begin{figure*}[!t]
    \centering
    \setlength{\abovecaptionskip}{0.0cm}
    \includegraphics[width=0.86\linewidth]{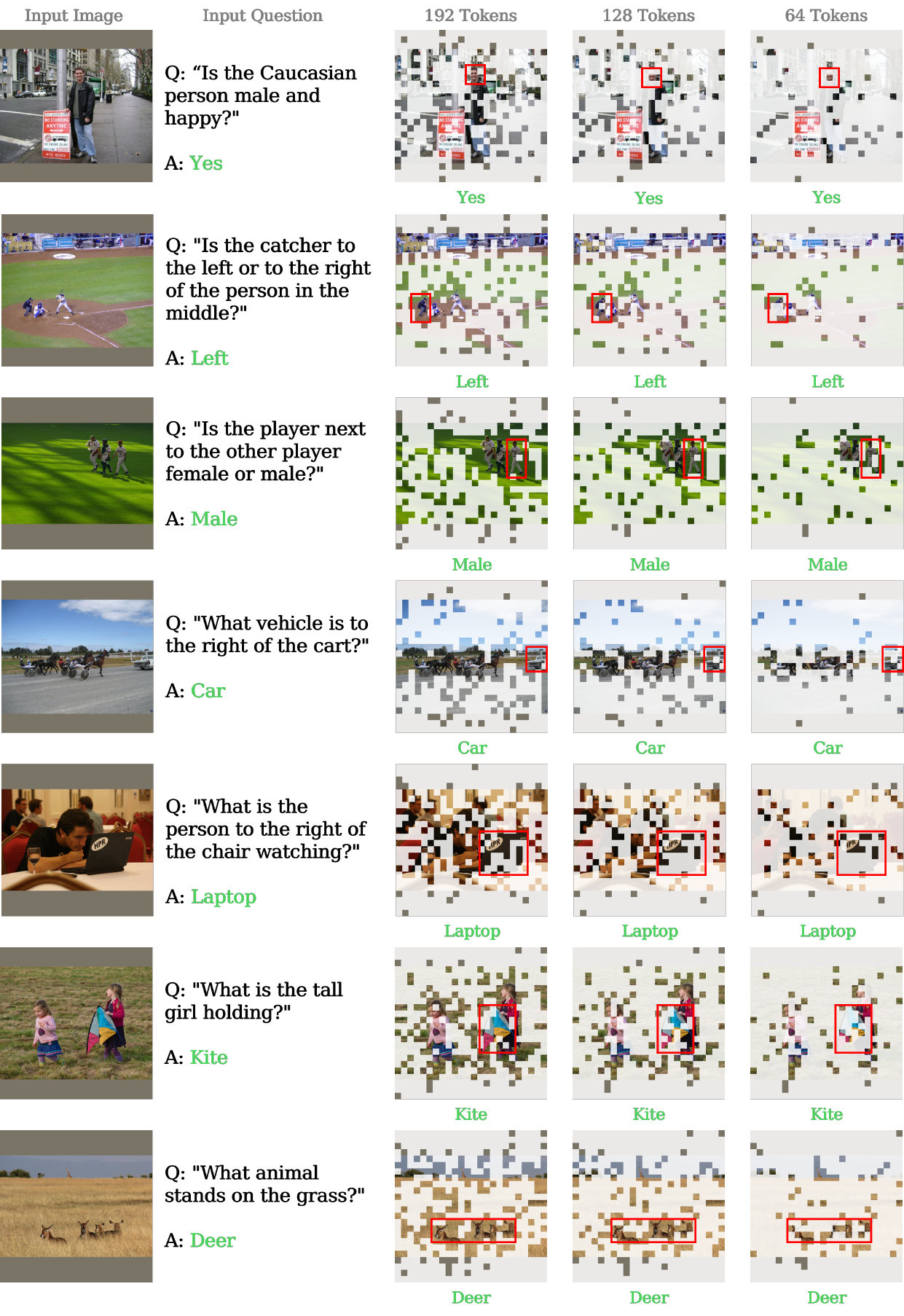}
\caption{
Additional success examples demonstrating that \ours preserves key visual cues across attributes, spatial relations, and object identification, enabling accurate predictions even under aggressive pruning.
}
    \label{fig:app:MoreCase}
\end{figure*}

\section{Additional Qualitative Results}
\label{appendix:visual}
\paragraph{Sensitivity vs. Attention vs. Diversity.}

Fig.~\ref{fig:appdix_visualization_att} compares token selection patterns driven by different token importance strategies. Attention-based methods exhibit clear limitations: text-visual (T2V) attention often suffers from positional bias (\eg disproportionately focusing on lower image regions, leading to the omission of key contextual cues), while visual-visual (V2V) attention tends to form redundant clusters. In contrast, diversity-based pruning treats all tokens equally, lacking semantic focus, whereas sensitivity-based selection captures key semantics but may result in spatial clustering. \ours\ addresses these shortcomings by unifying sensitivity and diversity. This ensures that selected tokens are both semantically informative and spatially distributed, maintaining robust coverage of task-relevant regions even under an aggressive 64-token budget.

\paragraph{More Visualization Examples.}

Fig.~\ref{fig:appendix_visualization_sota} visualizes the pruning masks and QA results. 
VisionZip prioritizes visual saliency, often focusing on the high-contrast kitchen background while missing the foreground ``Bench" across all ratios.
DivPrune enforces diversity but compromises object integrity (\eg it sparsifies ``Dog"-related tokens, leading to a Bear" hallucination).
ZOO-Prune overcomes these limitations by using sensitivity to anchor dense token clusters on semantic targets. It effectively preserves the visual cues of key objects, ensuring correct predictions even when 88\% of tokens are removed. This localization capability is further corroborated by Fig.~\ref{fig:app:MoreCase}, demonstrating that \ours\ consistently retains semantically critical tokens.

\paragraph{Failure Case Examples.}
Fig.~\ref{fig:app:FailedCase} illustrates representative failure cases under aggressive pruning, where \ours\ produces semantically close but inexact predictions. These errors reveal two limitations: difficulty in capturing fine-grained attributes for distinguishing closely related concepts, and vulnerability to visual clutter with multiple interfering objects. While our method preserves high-level semantic integrity at 64 tokens, extreme reduction inevitably compromises nuanced visual cues required for precise recognition in complex scenes.


\section{Discussion and Future Work}
\label{appendix:Future_Work}


Although \ours achieves strong gains across vision–language reasoning tasks, several avenues remain open.
First, our evaluation focuses on encoder–decoder VLMs such as LLaVA-NeXT. Extending \ours to broader architectures, including emerging Omni-style unified models, may reveal different sensitivity patterns and grounding behaviors. Second, the method is currently image-centric. Applying \ours to video, 3D scenes, or egocentric data will require reconsidering token selection under temporal and geometric structure. Finally, integrating \ours into Vision-Language-Action agents is a promising direction, as adaptive preservation of task-critical visual cues may improve long-horizon stability and reduce cascading errors. In sum, extending \ours toward Omni-style models, richer modalities, and interactive agents presents a promising direction for broadening the scope and impact of test-time visual refocusing.

\end{document}